\documentclass[10pt,journal,compsoc]{IEEEtran}
\ifCLASSOPTIONcompsoc
  \usepackage[nocompress]{cite}
\else
  \usepackage{cite}
\fi

\usepackage{times}
\usepackage[mathletters]{ucs}
\usepackage[utf8x]{inputenc}
\usepackage{epsfig}
\usepackage{amsmath}
\usepackage{bm}
\usepackage{amssymb}
\usepackage{mathabx}

\usepackage{mathrsfs}
\usepackage{caption}
\usepackage{pifont}
\usepackage{wasysym}
\usepackage{array}
\usepackage{caption}
\usepackage{color}
\usepackage{xcolor}
\usepackage{colortbl}
\usepackage{hyperref}
\usepackage{wrapfig}
\usepackage{multicol}
\usepackage{hhline}
\usepackage{booktabs} 
\usepackage{multirow}
\usepackage{color}
\usepackage{xcolor}
\usepackage{colortbl}
\usepackage{wrapfig}
\newcommand{\xmark}{\ding{55}}%
\newcommand{\eg}{\textit{e}.\textit{g}.}
\newcommand{\etal}{\textit{et al}.}
\newcommand{\ie}{\textit{i}.\textit{e}.}
\newcommand{\aka}{\textit{a.k.a.}}
\usepackage{multirow}
\usepackage{makecell, multirow, tabularx}
\usepackage{adjustbox}
\usepackage{tabularx}
\usepackage{xspace}
\usepackage{booktabs}
\usepackage{xurl}
\definecolor{hollywoodcerise}{rgb}{0.96, 0.0, 0.63}
\definecolor{lasallegreen}{rgb}{0.03, 0.47, 0.19}
\definecolor{hanpurple}{rgb}{0.32, 0.09, 0.98}
\definecolor{green(pigment)}{rgb}{0.0, 0.65, 0.31}

\usepackage{hyperref}
\hypersetup{colorlinks,linkcolor={red},citecolor={hollywoodcerise},urlcolor={red}}  

\usepackage{pdfpages} 
\usepackage{pgffor} 

\makeatletter
\AtBeginDocument{\let\LS@rot\@undefined}
\makeatother

\def\supplementfilename{supplement}

\pdfximage{\supplementfilename.pdf}
\def\numbersupplementpages{\the\pdflastximagepages}

\newif\ifarXiv
\arXivtrue 

\ifCLASSINFOpdf
\else
\fi
\begin{document}
%
\title{Deep Learning for Omnidirectional Vision: A Survey and New Perspectives}
%
%
%
%

\author{Hao~Ai$^*$, Zidong~Cao$^*$, Jinjing~Zhu, Haotian~Bai, Yucheng~Chen, and~ Lin~Wang 
\IEEEcompsocitemizethanks{
\IEEEcompsocthanksitem H. Ai and Z. Cao, J. Zhu, H. Bai, Y. Chen are with the Artificial Intelligence Thrust, The Hong Kong University of Science and Technology (HKUST), Guangzhou, China. E-mail: \{haoai, zidongcao, jinjingzhu, haotianbai, yuchengchen\}@ust.hk.
\IEEEcompsocthanksitem L. Wang is with the Artificial Intelligence Thrust, HKUST, Guangzhou, and Dept. of Computer Science and Engineering, HKUST, Hong Kong SAR, China. E-mail: linwang@ust.hk}
\thanks{Manuscript received April 19, 2022; revised August 26, 2022.
\hfil\break($^*$Equal  contribution, Corresponding author: Lin Wang)}
}

%
%

\markboth{Journal of \LaTeX\ Class Files,~Vol.~14, No.~8, August~2015}%
{Shell \MakeLowercase{\textit{et al.}}: Bare Demo of IEEEtran.cls for Computer Society Journals}
%



\IEEEtitleabstractindextext{%
\begin{abstract}
Omnidirectional image (ODI) data is captured with a $360^\circ \times 180^\circ$ field-of-view, which is much wider than the pinhole cameras and contains richer spatial information than the conventional planar images. Accordingly, omnidirectional vision has attracted booming attention due to its more advantageous performance in numerous applications, such as autonomous driving and virtual reality. In recent years, the availability of customer-level $360^\circ$ cameras has made omnidirectional vision more popular, and the advance of deep learning (DL) has significantly sparked its research and applications.
This paper presents a systematic and comprehensive review and analysis of the recent progress in DL methods for omnidirectional vision. 
Our work covers four main contents: (i) An introduction to the principle of omnidirectional imaging, the convolution methods on the ODI, and datasets to highlight the differences and difficulties compared with the 2D planar image data; (ii) A structural and hierarchical taxonomy of the DL methods for omnidirectional vision; (iii) A summarization of the latest novel learning strategies and applications; (iv) An insightful discussion of the challenges and open problems by highlighting the potential research directions to trigger more research in the community. 
\end{abstract}

}

\maketitle

\IEEEdisplaynontitleabstractindextext

%
\IEEEpeerreviewmaketitle

\IEEEraisesectionheading{\section{Introduction}\label{sec:introduction}}
%
%
%
%
\IEEEPARstart{W}{ith} the rapid development of 3D technology and the pursuit of realistic visual experience, research interest in computer vision has gradually shifted from traditional 2D planar image data to omnidirectional image (ODI) data, also known as the 360$^\circ$ image, panoramic image, or spherical image data. 
ODI data captured by the $360^\circ$ cameras yields a $360^\circ \times 180^\circ$ field-of-view (FoV), which is much wider than the pinhole cameras; therefore, it can capture the entire surrounding environment by reflecting richer spatial information than the conventional planar images. Due to the immersive experience and complete view, ODI data has been widely applied to numerous applications, \eg, augmented reality(AR)$/$virtual reality (VR), autonomous driving, and robot navigation. 
In general, raw ODI data is represented as, \eg, the equirectangular projection (ERP) or cubemap projection (CP) to be consistent with the imaging pipelines~\cite{Pi2020ContentawareHE},~\cite{Jiang2021CubemapBasedPB}.
As a novel data domain,  ODI data has both domain-unique advantages (wide FoV of spherical imaging, rich geometric information, multiple projection types) and challenges (severe distortion in the ERP type, content discontinuities in the CP format). This renders the research on omnidirectional vision valuable yet challenging.


Recently, the availability of customer-level $360^\circ$ cameras has made omnidirectional vision more popular, and the advance in deep learning (DL) has significantly promoted its research and applications. In particular, as a data-driven technology, the continual release of public datasets, \eg,~SUN360~\cite{xiao2012recognizing}, Salient 360$!$~\cite{rai2017dataset}, Stanford2D3D~\cite{armeni2017joint}, Pano-AVQA~\cite{Yun2021PanoAVQAGA} and PanoContext~\cite{Zhang2014PanoContextAW}, have rapidly enabled the DL methods to accomplish remarkable breakthroughs and often achieve the state-of-the-art (SoTA) performances on various omnidirectional vision tasks. Moreover, various deep neural network (DNN) models have been developed based on diverse architectures, ranging from convolutional neural networks (CNNs)~\cite{He2016DeepRL}, recurrent neural networks (RNNs)~\cite{medsker2001recurrent}, generative adversarial networks (GANs)~\cite{goodfellow2014generative}, graph neural networks (GNNs)~\cite{scarselli2008graph}, to vision transformers (ViTs)~\cite{dosovitskiy2020image}. In general, SoTA-DL-methods focus on four major aspects: (I) convolutional filters used to extract features from the ODI data (omnidirectional video (ODV) can be considered as a temporal set of ODIs), (II) network design by considering the input numbers and projection types, (III) novel learning strategies, and (IV) practical applications.



\begin{figure*}[ht]
    \centering
    \includegraphics[width=.92\textwidth]{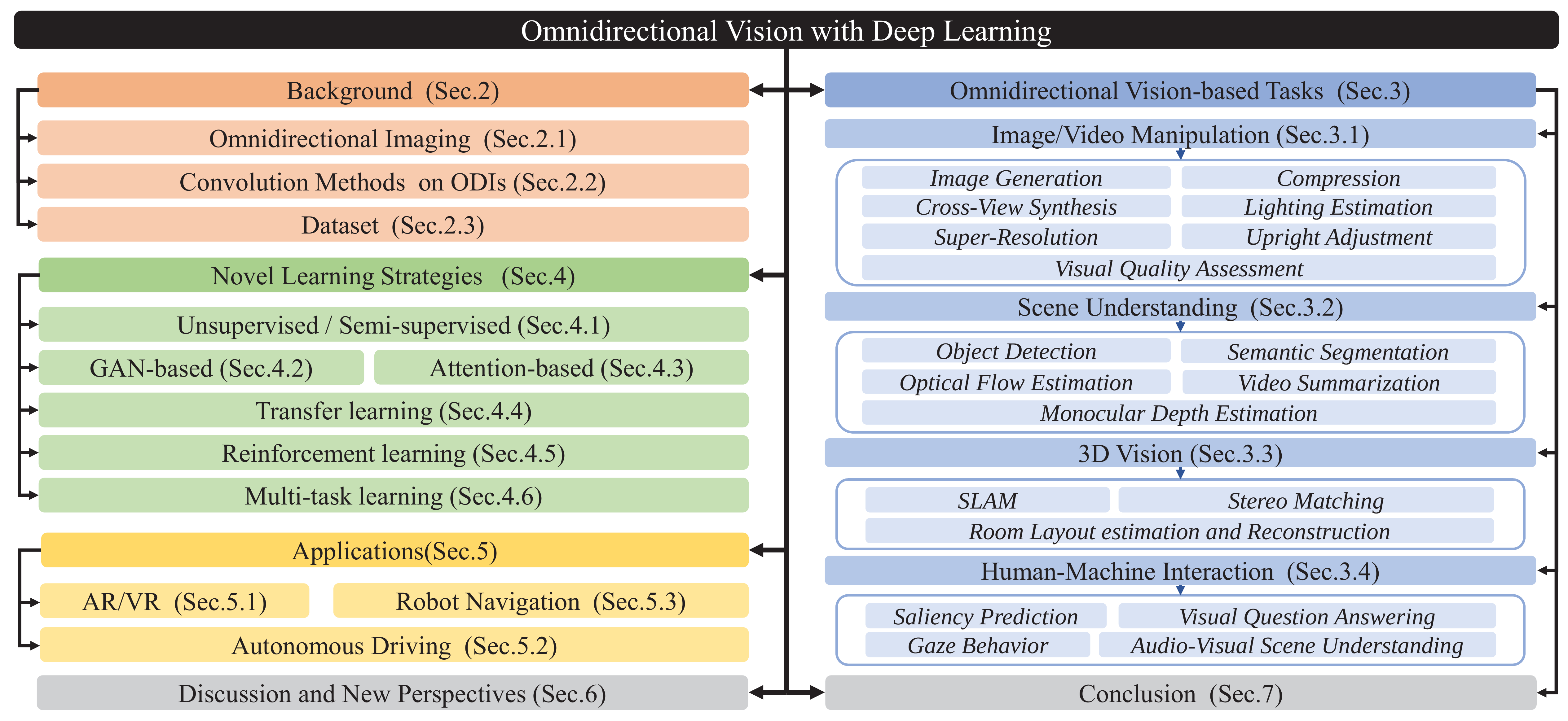}
    \vspace{-8pt}
    \caption{Hierarchical and structural taxonomy of omnidirectional vision with deep learning.}
    \label{fig:survey}
     \vspace{-11pt}
\end{figure*}

This paper presents a systematic and comprehensive review and analysis of the recent progress in DL methods for omnidirectional vision.
Previously, Zou~\etal~\cite{Zou2021ManhattanRL} only focused on the algorithms of reconstructing room layout from a single ODI based on the Manhattan assumption. Similarly, Silveira \etal~\cite{daSilveira20223DSG} merely reviewed recent 3D scene geometry recovery approaches based on the ODIs. Moreover, there exist some limited reviews of the FoV-adaptive video streaming methods~\cite{Zink2019Scalable3V},~\cite{Xu2020StateoftheArtI3}, especially on the topic of projection types, visual distortion problems, and efficient network structures. Recently, Chiariotti \etal~\cite{yaqoob2020survey} provided a more extensive review of the existing literature about ODV streaming systems. \textit{Unlike them, we highlight the importance of DL and probe the recent advances for omnidirectional vision, both methodically and comprehensively}. The structural and hierarchical taxonomy proposed in this study is shown in Fig.~\ref{fig:survey}.

In summary, the major contributions of this study can be summarized as follows: (\uppercase\expandafter{\romannumeral1}) To the best of our knowledge, this is the \textbf{first} survey to comprehensively review and analyze the DL methods for omnidirectional vision, including the omnidirectional imaging principle, representation learning, datasets, a taxonomy, and applications, to highlight the differences and difficulties with the 2D planner image data. (\uppercase\expandafter{\romannumeral2}) We summarize most, if not all but representative, published top-tier conference/journal works (over 200 papers) in the last five years and conduct an analytical study of recent trends of DL for omnidirectional vision, both hierarchically and structurally. Moreover, we offer insights into the discussion and challenge of each category. (\uppercase\expandafter{\romannumeral3}) We summarize the latest novel learning strategies and potential applications for omnidirectional vision. (\uppercase\expandafter{\romannumeral4}) As DL for omnidirectional vision is an active yet intricate research area, we provide insightful discussions of the challenges and open problems yet to be solved and propose the potential future directions to spur more in-depth research by the community. Meanwhile, we have summarized representative methods and their key strategies for some popular omnidirectional vision tasks in Table.~\ref{table:view synthesis}, Table.~\ref{table:segmentation}, Table.~\ref{table:depth}, Table.~\ref{table:room layout}, and Table.~\ref{table:saliency}. To provide a better intra-task comparison, we present some representative methods' quantitative and qualitative results on benchmark datasets and all statistics are derived from the original papers. \textit{Due to the lack of space, we show the experimental results in Sec.~2 of the suppl. material.} (\uppercase\expandafter{\romannumeral5}) We create an open-source repository that provides a taxonomy of all the mentioned works and code links. We will keep updating our open-source repository with new works in this area and hope it can shed light on future research. The repository link is \url{https://github.com/VLISLAB/360-DL-Survey}.
\begin{figure}[t!]
    \centering
    \includegraphics[width=0.5\linewidth]{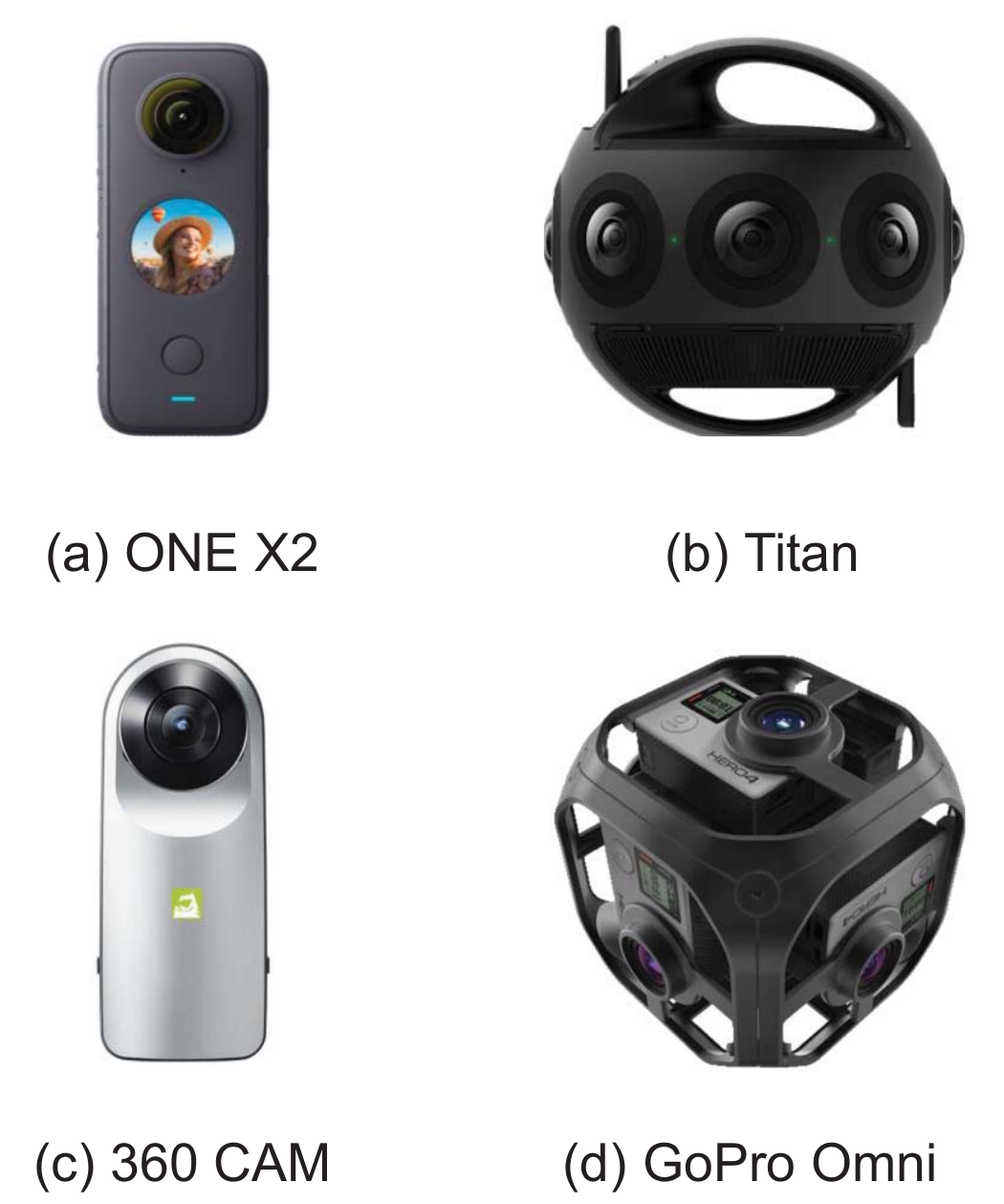}
     \vspace{-7pt}
    \caption{Examples of representative $360^\circ$ cameras.}
    \label{fig:camera}
    \vspace{-18pt}
\end{figure}
\begin{figure*}[htbp]
    \centering
    \includegraphics[width=.98\textwidth]{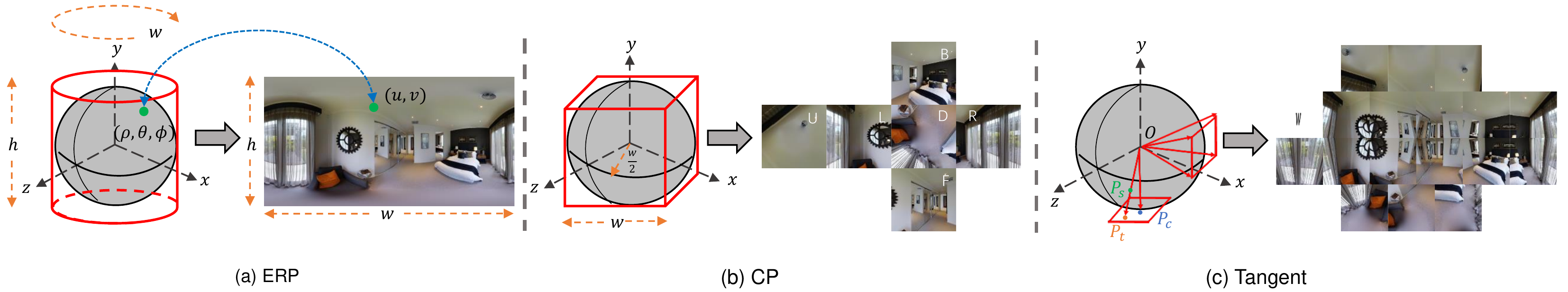} 
     \vspace{-6pt}
    \caption{Illustration of ERP, CP and tangent representation types.}
    \label{fig:imaging}
    \vspace{-12pt}
\end{figure*}

The rest of the paper is organized as follows. In Sec.~\ref{sec2}, we introduce the imaging principle of ODI, convolution methods for omnidirectional vision, and some representative datasets. Sec.~\ref{sec3} introduces the existing DL approaches for various tasks and provides taxonomies to categorize the relevant papers. Sec.~\ref{sec4} covers novel learning paradigms for the tasks in omnidirectional vision, \eg, unsupervised learning, transfer learning, and reinforcement learning. Sec.~\ref{sec5} then scrutinizes the applications, followed by Sec.~\ref{sec6}, where we discuss open problems and future directions. Finally, we conclude this paper in Sec.~\ref{sec7}.
\section{Background}
\label{sec2}
\subsection{Omnidirectional Imaging}
\label{sec2.1}

\subsubsection{Acquisition}
\label{sec2.1.1}
A normal camera has an FoV less than $180^{\circ}$ and thus captures view at most a hemisphere. However, an ideal $360^\circ$ camera can capture lights falling on the focal point from all directions, making the projection plane a whole spherical surface. In practice, most $360^\circ$ cameras can not achieve it, which excludes top and bottom regions due to dead angles\footnote{\url{https://en.wikipedia.org/wiki/Omnidirectional_(360-degree)_camera}}. According to the number of lenses, $360^\circ$ cameras can be categorized into three types: (i) Cameras with one fisheye lens, which is impossible to cover the whole spherical surface. However, if the intrinsic and extrinsic parameters are known, an ODI can be achieved by projecting multiple images into a sphere and stitching them together; (ii) Cameras with dual fisheye lenses located at opposite positions, each of which covers over $180^{\circ}$ FoV, such as Insta360 ONE\footnote{\url{https://www.insta360.com/product/insta360-one}} and LG 360 CAM\footnote{\url{https://www.lg.com/sg/lg-friends/lg-360-CAM}\label{lg-360-CAM}}. This type of $360^\circ$ cameras have minimum demand for lenses, which are cheap and convenient, favoured  by industries and customers. Images from the two cameras are then stitched together to obtain an omnidirectional image, but the stitching process might lead to edge blurring; (iii) Cameras with more than two lenses, such as Titan (eight lenses)\footnote{\url{https://www.insta360.com/product/insta360-titan/}\label{titan}}. In addition, GoPro Omni\footnote{\url{https://gopro.com/en/us/news/omni-is-here}\label{gopro}} is the first camera rig to place six regular cameras onto six faces of a cube and its synthesized results have higher precision and less blur in edges. This type of $360^{\circ}$ cameras are professional-level.
\subsubsection{Spherical Imaging}
\label{sec2.1.2}
We first define the spherical coordinate $(\theta, \phi, \rho)$, where $\theta \in (0, 2\pi)$ ,$\phi \in (0, \pi)$, and $\rho$ represent the latitude, longitude, and radius of the sphere, respectively. We also define the Cartesian coordinate $(x,y,z)$. The transformation between spherical coordinate and Cartesian coordinate can be formulated as follows~\cite{zioulis2019spherical}:  
\vspace{-2pt}
\begin{equation}
    \begin{array}{|c|}
         \rho  \\
         \theta \\
         \phi \\
    \end{array} = 
    \begin{array}{|c|}
         (x^2+y^2+z^2)^{1/2} \\
         \arctan(x/z) \\
         \arccos(y/\rho) \\
    \end{array} \ , \ 
    \begin{array}{|c|}
         x  \\
         y \\
         z \\
    \end{array} = 
    \begin{array}{|c|}
         \rho\sin(\theta)\sin(\phi) \\
         \rho\cos(\phi) \\
         \rho\cos(\theta)\sin(\phi) \\
    \end{array}.
    \label{eq.1}
\end{equation}
\vspace{-2pt}

\noindent \textbf{Equirectangular Projection} (\textbf{ERP})\footnote{\url{https://en.wikipedia.org/wiki/Equirectangular_projection}\label{erp}} is a representation by uniformly sampling grids from the spherical surface, as shown in Fig.~\ref{fig:imaging}(a). The horizontal unit angle is $\vartheta=2\pi/w$ and the vertical unit angle is $\varphi=\pi/h$. In particular, if the horizontal and vertical unit angle are equal, the width $w$ is twice of height $h$. In a word, each pixel coordinate $(u,v)$ in ERP can be mapped to the spherical coordinate $(\theta,\phi)=(u \cdot  \vartheta,v\cdot \varphi)$ and vice versa.
\noindent \textbf{Cubemap Projection} (\textbf{CP}) projects the spherical surface to six cube faces with $90^{\circ}$ FoV, equal-side length $w$, and focal length $\frac{w}{2}$, as shown in Fig.~\ref{fig:imaging}(b). We denote the cube faces as $f_i$, $i \in \{B,D,F,L,R,U\}$, representing back, down, front, left, right, and up, respectively. By setting the cube center as the origin, the extrinsic matrix of each face can be simplified into $90^\circ$ or $180^\circ$ rotation matrix and zero translation matrix~\cite{wang2020bifuse}. Given a pixel on the plane $f_i$, we transform $f_i$ to the front plane (identical to the Cartesian coordinates) and calculate $(\theta, \phi)$ with Eq.~\ref{eq.1}.

\begin{figure*}[htbp]
    \centering
    \includegraphics[width=.97\textwidth]{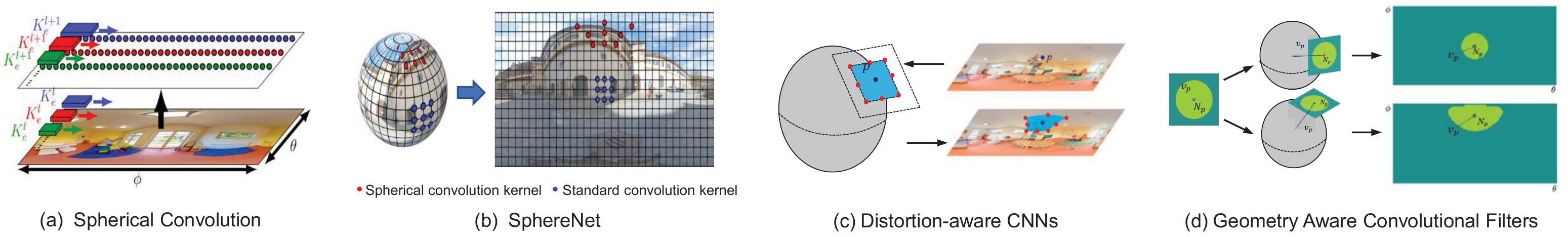}
    \caption{An illustration of ERP-based convolution filters on ODIs. (a), (b), (c) and (d) are originally shown in~\cite{Su2017LearningSC},~\cite{coors2018spherenet},~\cite{zhao2018distortion} and~\cite{Khasanova2019GeometryAC}.}
    \label{fig:planarproj1}
    \vspace{-0.3cm}
\end{figure*}
\noindent \textbf{Tangent Projection} is the gnomonic projection~\cite{o1962introduction}, a non-conformal projection from points $P_s$ on the sphere surface with the sphere center $O$ to points $P_t$ in a tangent plane with center $P_c$~\footnote{\url{https://mathworld.wolfram.com/GnomonicProjection.html}}, as shown in Fig.~\ref{fig:imaging}(c). For a pixel on the ERP image $P_{e}(u_e,v_e)$, we first calculate its corresponding point $P_{s}(\theta=u_{e} \cdot \vartheta, \phi=v_{e} \cdot \varphi)$ on the unit sphere, following the transformation in \textbf{ERP} format. The projection from $P_{s}(\theta,\phi)$ to $P_t(u_t,v_t)$ is defined as:

\begin{equation}
\begin{split}
&    u_t=\frac{\cos(\phi)\sin(\theta-\theta_c)}{\cos{c}}, \\
&    y_t=\frac{\cos(\phi_c)\sin(\phi)-\sin(\phi_c)\cos(\phi)\cos(\theta-\theta_c)}{\cos(c)}, \\
&   \cos(c)=\sin(\phi_c)\sin(\phi)+\cos(\phi_c)\cos(\phi)\cos(\theta-\theta_c),
\end{split}
\label{eq.2}
\end{equation}

\noindent where $(\theta_c,\phi_c)$ is the spherical coordinate of the tangent plane center $P_c$, and $(u_t,v_t)$ is the intersection coordinate of the   tangent plane and the extension line of $\overrightarrow{OP_s}$. The inverse transformations are formulated as:

\begin{equation}
\begin{split}
& \theta = \theta_c + \tan^{-1}(\frac{u_t\sin(c)}{\gamma\cos(\phi_c)\cos(c)-v_t\sin(\phi_c)\sin(c)}), \\
& \phi=\sin^{-1}(\cos(c)\sin(\phi_c)+\frac{1}{\gamma}v_t\sin(c)\cos(\phi_c)),
\end{split}
\label{eq.3}
\end{equation}
where $\gamma=\sqrt{u_{t}^{2}+v_{t}^{2}}$ and $c=\tan^{-1}\gamma$. With Eqs. \ref{eq.2} and \ref{eq.3}, we can build one-to-one forward and inverse mapping functions between the spherical coordinates and pixels on the tangent images~\cite{li2022omnifusion}.

\noindent \textbf{Icosahedron} approximates a sphere surface through a Platonic solid~\cite{eder2020tangent}. Compared with ERP and CP, icosahedron projection has resolved the spherical distortion well. While some practical applications need less distortion representations, we can increase the number of subdivisions to further mitigate the spherical distortion. Specifically, each face in an icosahedron can be subdivided into four smaller faces to achieve higher resolution and less distortion~\cite{eder2020tangent}. There exist some CNNs that are specifically designed to process an icosahedron~\cite{Lee2019SpherePHDAC, yoon2021spheresr}. It is noteworthy that the choice of subdivision 
degree needs to achieve a trade-off between efficiency and accuracy.


\noindent \textbf{Other projections.} For CP, different sampling locations on the cube faces decide different spatial sampling rates, resulting in the distortions. To address this problem, Equi-Angular Cubemap (EAC) projection~\footnote{\url{https://blog.google/products/google-ar-vr/bringing-pixels-front-and-center-vr-video/}} is proposed to keep the sampling uniform. Besides, some projections can transform the spherical surface into non-spatial domains, \eg, 3D rotation group (SO3)~\cite{cohen2018spherical} and spherical Fourier transformation (SFT)~\cite{cruz2012scale}. 

\vspace{-5pt}
\subsubsection{Spherical Stereo}
\label{sec2.1.3}
Spherical stereo is about two viewpoints displaced with a known horizontal or vertical baseline~\cite{zioulis2019spherical}. Due to the spherical projection, spherical stereo is more irregular than stereo with traditional pinhole cameras. According to Eq.~\ref{eq.1}, we define the baseline as $\textbf{b}=(\delta x, \delta y, \delta z)$, and the derivative correspondence between the spherical coordinates and Cartesian coordinates can be formulated as follows:
\begin{equation}
    \begin{array}{|c|}
        \delta_\rho  \\
        \delta_\theta \\
        \delta_\phi \\
    \end{array} = 
    \begin{array}{|c c c|}
         \sin(\theta)\sin(\phi) & \cos(\phi) & \cos(\theta)\sin(\phi) \\
         \frac{\cos(\theta)}{\rho\sin(\phi)} & 0 & \frac{-\sin(\theta)}{\rho\sin(\phi)} \\
         \frac{\sin(\theta)\cos(\phi)}{\rho} & \frac{-\sin(\phi)}{\rho} & \frac{\cos(\theta)\cos(\phi)}{\rho} \\
    \end{array} \
     \begin{array}{|c|}
        \delta x  \\
        \delta y \\
        \delta z \\
    \end{array}.
    \label{eq.4}
\end{equation}

In Eq.~\ref{eq.4}, $(\delta_\theta, \delta_\phi)$ represents the angular differences in the spherical coordinates $(\theta, \phi,\rho)$. According to Eq. \ref{eq.4}, we can find that for the vertical baseline $\textbf{b}_v=(0,\delta y,0)$, there is no difference in $\theta$, which is simpler. However, for horizontal baseline $\textbf{b}_h=(\delta x,0,0)$, differences occur in both angles $\theta$ and $\phi$.

\vspace{-5pt}
\subsection{Convolution Methods on ODI}
\label{sec2.2}
As the natural projection surface of an ODI is a sphere, standard CNNs are less capable of processing the inherent distortions when the spherical image is projected back to a plane. Numerous CNN-based methods have been proposed to enhance the extraction of "unbiased" information from spherical images. These methods can be classified into two prevailing categories: (i) Applying 2D convolution filters on planar projections; (ii) Directly leveraging spherical convolution filters in the spherical domain. In this subsection, we analyze these methods in detail.

\vspace{-2pt}
\subsubsection{Planar Projection-based Convolution}
\label{sec2.2.1}
As the most common sphere-to-plane projection, ERP introduces severe distortions, especially at the poles. Considering it provides global information and takes less computation cost, Su \etal~\cite{Su2017LearningSC} proposed a representative method named Spherical Convolution, which leverages regular convolution filters with the adaptive kernel size according to the spherical coordinates. However, as shown in Fig.~\ref{fig:planarproj1}(a), the regular convolution weights are only shared along each row and can not be trained from scratch.
Inspired by Spherical Convolution, SphereNet~\cite{coors2018spherenet} proposes another typical method that processes the ERP by directly adjusting the sampling grid locations of convolution filters to achieve the distortion invariance and can be trained end-to-end, as depicted in Fig.~\ref{fig:planarproj1}(b). This is conceptually similar to those in~\cite{zhao2018distortion}, \cite{Khasanova2019GeometryAC}, as shown in Fig.~\ref{fig:planarproj1}(c) and (d). In particular, before ODIs are widely applied, Cohen \etal~\cite{cohen2018spherical} have discussed the spatially varying distortions introduced by ERP and proposed a rotation-invariant spherical CNN approach to learn an SO3 representation. By contrast, KTN~\cite{Su2019KernelTN,Su2021LearningSC} learns a transfer function to achieve that the convolution kernel, which is learnt from the conventional planar images, can be directly applied on ERP without retraining. In~\cite{Frossard2017GraphBasedCO}, the ERP is represented as a weighted graph, and a novel graph construction method is introduced by incorporating the geometry of the omnidirectional cameras into the graph structure to mitigate the distortions.~\cite{wang2020bifuse, reyarea2021360monodepth} focused on directly applying traditional 2D CNNs on CP and tangent projection, which are distortion-less.
\begin{figure}[t!]
    \centering
    \includegraphics[width=0.77\linewidth]{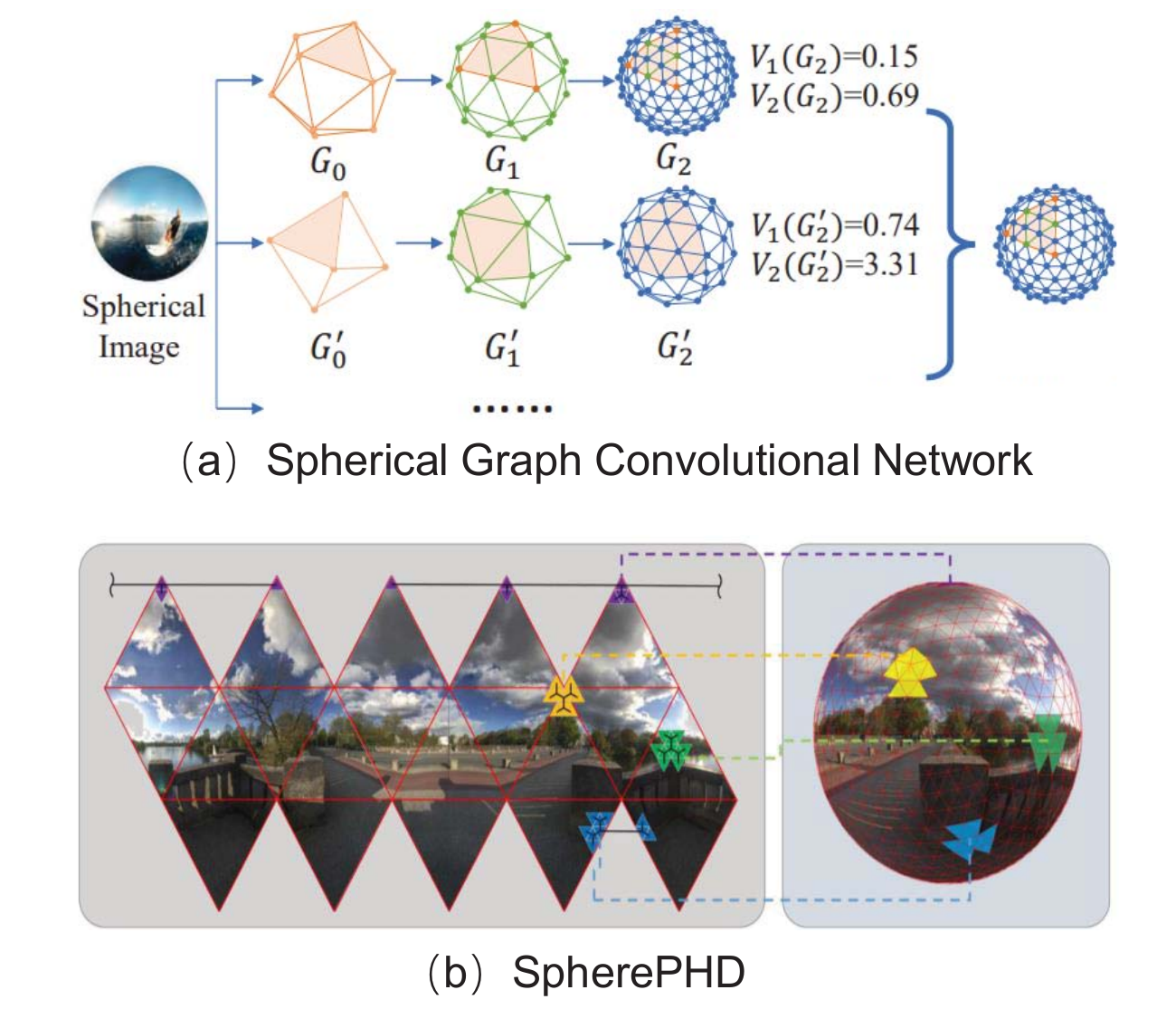}
    \vspace{-8pt}
    \caption{The two representative spherical convolution approaches. (a) and (b) are originally shown in~\cite{Yang2020RotationEG} and~\cite{Lee2019SpherePHDAC}.}
    \label{fig:planarproj2}
    \vspace{-12pt}
\end{figure}
\vspace{-2pt}
\subsubsection{Spherical Convolution}
\label{sec2.2.2}
Some methods have explored the special convolution filters in the spherical domain. Esteves \etal~\cite{esteves2018learning} proposed the first spherical CNN architecture, which considers the convolution filters in the spherical harmonic domain, to address the problem of 3D rotation equivariance in standard CNNs. Unlike~\cite{esteves2018learning}, Yang \etal~\cite{Yang2020RotationEG} proposed a representative framework to map spherical images into the rotation-equivariant representations based on the geometry of spherical surfaces. As shown in Fig.~\ref{fig:planarproj2}(a), SGCN~\cite{Yang2020RotationEG} represents the input spherical image as a graph based on the GICOPix~\cite{Yang2020RotationEG}. Moreover, it explores the isometric transformation equivariance of the graph through GCN layers. A similar strategy is proposed in~\cite{cohen2019gauge} and~\cite{shakerinava2021equivariant}. In \cite{cohen2019gauge}, the gauge equivariant CNNs are proposed to learn spherical representations from the icosahedron. By contrast,  Shakerinava \etal~\cite{shakerinava2021equivariant} extended the icosahedron to all the pixelizations of platonic solids and generalized the gauge equivariant CNNs on the pixelized spheres. Due to a trade-off between efficiency and rotation equivariance, DeepSphere~\cite{Defferrard2020DeepSphere} models the sampled sphere as a graph of connected pixels and designs a novel graph convolution network (GCN) to balance the computational efficiency and sampling flexibility by adjusting the neighboring  pixel numbers of the pixels on the graph. Compared with the methods above, another representative ODI representation is proposed in SpherePHD~\cite{Lee2019SpherePHDAC}. As shown in Fig.~\ref{fig:planarproj2}(b), SpherePHD represents the spherical image as the spherical polyhedron and provides specific convolution and pooling methods.

\vspace{-5pt}
\subsection{Dataset}
\label{sec2.3}
\begin{table*}[htbp]
\setlength{\abovecaptionskip}{-0.1cm}
\caption{Summary of ODI image and video datasets. N/A indicates `not available' and GT indicates `ground truth'.}
\begin{center}
\renewcommand\arraystretch{1.2}
    \begin{tabularx}{0.8\textwidth}{c|c|c|c|c|m{3.5cm}<{\centering}}
    \hline
    Dataset & Size & Data Type & Resolution  & GT & Purpose \\
    \hline
    \hline
    Stanford2D3D~\cite{armeni2017joint} & 70496 RGB+1413 ERP images & Real& 1080 $\times$ 1080 & $\checkmark$ &Object Detection, Scene Uderstanding  \\ \hline
    Structured3D~\cite{zheng2020structured3d} & 196k images & Synthetic&512 $\times$ 1024&\ding{55} & Object Detection, Scene Understanding, Image Synthesis, 3D Modeling\\ \hline
    SUNCG~\cite{song2017semantic} & 45622 scenes & Synthetic & N/A &\ding{55} & Depth Estimation \\ 
    \hline
    \hline
    360-Sport~\cite{hu2017deep} & 342 360$^\circ$ videos & Real & N/A & $\checkmark$ & Visual Pilot \\ \hline
    Wild-360~\cite{cheng2018cube} & 85 360$^\circ$ videos & Real & N/A & $\checkmark$ & Video Saliency \\ 
\hline
\end{tabularx}
\end{center}
\label{table:dataset}
\vspace{-10pt}
\end{table*}
The performance of the DL-based approaches is closely related to the qualities and quantities of the datasets. With the development of spherical imaging devices, a large number of ODI and ODV datasets are publicly available for various vision tasks. Especially, most ODV data is collected from public video sharing platforms like Vimeo and Youtube. In Table.~\ref{table:dataset}, we list some representative ODI and ODV datasets used for different purposes and we also show their properties, \eg, size, resolution, data source. \textit{Complete summary of datasets can be found in the suppl. material.}  According to the data source, there are two categories of datasets: real-world datasets and synthetic datasets. Most real-world datasets only provide images of 2D projection modality and are applied to some specific task. However, Stanford2D3D~\cite{armeni2017joint} contains three modalities, including 2D, 2.5D, that are suitable for cross-modal learning. Moreover, some datasets are selected from the existing ones, such as PanoContext~\cite{Zhang2014PanoContextAW} collected from SUN360~\cite{xiao2012recognizing}. For the synthetic datasets, images are complete and high-quality without natural noise, and the annotations are easier to obtain than that in the real-world scenes. For instance, SUNCG~\cite{song2017semantic} is created via the Plannar5D platform, and all the 3D scenes are composed of individually labeled 3D object meshes. Structured3D~\cite{zheng2020structured3d} and OmniFlow~\cite{Seidel2021OmniFlowHO} utilize the rendering engine to generate photo-realistic images containing 3D structure annotations and corresponding optical flows. Similar to real-world datasets, there are also some datasets, \eg, omni-SYNTHIA~\cite{zhang2019orientation}, extracted from the large synthetic ones for specific tasks.

\vspace{-3pt}
\section{Omnidirectional Vision Tasks}
\label{sec3}

\subsection{Image/Video Manipulation}
\label{sec3.1}

\subsubsection{Image Generation}
\label{sec3.1.1}
\noindent \textbf{Insight:} \textit{Image generation aims to restore or synthesize the complete and clean ODI data from the partial or noisy data. }

For image generation on ODI, there exist four popular research directions: (i) panoramic depth map completion; (ii) ODI completion; (iii) panoramic semantic map completion; (iv) view synthesis on ODI. In this subsection, we provide a comprehensive analysis of some representative works.

\noindent \textbf{Depth Completion:} 
Due to the scarcity of real-world sparse-to-dense panoramic depth maps, this task mainly utilizes simulation techniques to generate artificially sparse depth maps as the training data. Liu \etal~\cite{liu2022cross} proposed a representative two-stage framework to achieve panoramic depth completion. In the first stage, a spherical normalized convolution network is proposed to predict the initial dense depth maps and confidence maps from the sparse depth inputs. Then the output of the first stage is combined with corresponding ODIs to generate the final panoramic dense depth maps through a cross-modal depth completion network. Especially, BIPS~\cite{oh2021bips} proposes a GAN framework to synthesize RGB-D indoor panoramas from the limited input information about a scene captured by the camera and depth sensors in arbitrary configurations. However, BIPS ignores a large distribution gap between synthesized and real LIDAR scanners, which could be better addressed with domain adaptation techniques.

\noindent \textbf{ODI Completion:} It aims to fill in missing areas to generate complete and plausible ODIs. Considering the high degree of freedom involved in generating an ODI from a single limited FoV image, Hara \etal~\cite{hara2020spherical} leveraged a fundamental property of the spherical structure, scene symmetry, to control the degree of freedom and improve the plausibility of the generated ODI. On the opposite of~\cite{hara2020spherical}, Akimoto \etal~\cite{akimoto2022diverse} proposed a transformer-based framework to synthesize the ODIs with arbitrary resolution from a fixed limited FoV image and encouraged the diversity of synthesized ODIs. In addition, Sumantri \etal~\cite{sumantri2020360} proposed a first pipeline to reconstruct the ODIs from a set of unknown FoV images without any overlap, including two steps: (i) FoV estimation of input images relative to the panorama; (ii) ODI synthesis with the input images and estimated FoVs.

\noindent \textbf{Semantic Scene Completion (SSC):} It aims to reconstruct the indoor scenes with both the occupancy and semantic labels of the whole room. Existing works, \eg,~\cite{roldao20213d}, are mostly based on the RGB-D data and LiDAR scanners. As the first work to accomplish the SSC task using the ODI data,~\cite{dourado2020semantic} used only a single ODI and its corresponding depth map as the input and generates a voxel grid from the input panoramic depth map. This voxel grid is partitioned into eight overlapping views, and each partitioned grid, representing a single view of a regular RGB-D sensor, is submitted to the 3D CNN model~\cite{dai2018scancomplete}, pre-trained on the standard 2.5D synthetic RGB-D data. These partial inferences are aligned and ensembled to obtain the final result.

\begin{table*}[htbp]
\setlength{\abovecaptionskip}{-0.1cm}
\caption{Cross-view synthesis and geo-localization by some representative methods.}
\begin{center}
\renewcommand\arraystretch{1.5}
    \begin{tabularx}{0.97\textwidth}{c|c|c|c|c|X<{\centering}}
    \hline
    Method & Publication & Input & View synthesis & Localization & Highlight  \\
    \hline
    \hline
    Lu \cite{lu2020geometry} & CVPR'20 & Image & $\checkmark$   & \ding{55} & Utilizing depth and semantics \\
    Li \cite{li2021sat2vid}    & ICCV'21  & Video & $\checkmark$   &\ding{55} & 3D point cloud representation with depth and semantics \\
    Zhai \cite{zhai2017predicting}  & CVPR'17 & Image & $\checkmark$  & $\checkmark$ & Pretraining semantic segmentation task with transfer learning \\
    Regmi \cite{regmi2019bridging}  & ICCV'19 & Image & $\checkmark$  & $\checkmark$ & Two stage training: Satellite-view synthesis and feature matching \\
    Toker \cite{toker2021coming} & CVPR'21 & Image & $\checkmark$  & $\checkmark$ & End-to-end training for view synthesis and feature matching  \\
    Shi \cite{shi2019spatial}  & NIPS'19 & Image &\ding{55} & $\checkmark$ & Polar transform  \\
    Zhu \cite{zhu2022transgeo}  & CVPR'22 & Image & \ding{55} & $\checkmark$ & Attention-based transformer and remove uninformative patches \\
    Shi \cite{shi2020looking}  & CVPR'20 & Image &\ding{55} & $\checkmark$ & Adding orientation estimation during localization \\
    Zhu \cite{zhu2021vigor}  & CVPR'21 & Image &\ding{55} & $\checkmark$ & Proposing that multiple satellite images can cover one ground image \\

\hline
\end{tabularx}
\end{center}
\label{table:view synthesis}
\vspace{-0.3cm}
\end{table*}

\noindent \textbf{View Synthesis:} View synthesis aims to generate ODIs from unknown viewpoints. 
OmniNeRF, proposed by Hsu \etal~\cite{hsu2021moving}, is the first and representative learning approach for panoramic view synthesis. To generate a novel view ODI, it first projects an ODI to the 3D domain with an auxiliary depth map and a derived gradient image, and then translates the view position to re-project the 3D coordinates to 2D space. The neural radiance fields (NeRF)~\cite{mildenhall2020nerf} is used to learn the pixel-based representations and solve the information missing problem caused by viewpoint translation. A similar strategy, proposed by~\cite{hara2022enhancement}, leverages a conditional generator to synthesize the novel view. With video as the input, Pathdreamer~\cite{koh2021pathdreamer} designs a hierarchical architecture to conduct the non-observed view synthesis from one previous observation and the trajectory of future viewpoints.

\vspace{-3pt}
\subsubsection{Cross-view Synthesis and Geo-localization}
\label{sec3.1.2}
\noindent \textbf{Insight:} \textit{Cross-view synthesis aims to synthesize ground-view ODIs from the satellite-view images while geo-localization aims to match the ground-view ODIs and satellite-view images to determine their relations.} 

Ground-view, \aka, street-view images are usually panoramic to provide complete surrounding information, while satellite views are planar images captured to cover almost every corner of the world. There exist a few methods to synthesize ground-view images from satellite-view images. Lu \etal~\cite{lu2020geometry} proposed a representative work including three stages: satellite stage, geo-transformation stage, and street-view stage. The satellite stage predicts depth maps and segmentation maps from satellite images. The geo-transformation stage transforms the output of the satellite stage into the panoramas. Finally, the street-view stage predicts the street-view panoramas from the segmentation maps via a GAN. Sat2Vid~\cite{li2021sat2vid}, the first work for cross-view video synthesis, also employs three stages to generate street-view ODVs using voxel grids with semantics and depth cues transformed from satellite images with trajectory. This is conceptually similar to that in~\cite{lu2020geometry}.

In general, the framework for geo-localization consists of two modules: cross-synthesis module and retrieval module. Shi \etal~\cite{shi2019spatial} proposed a representative contrastive learning  pipeline to calculate the distance between the ground-view ODIs and satellite-view images in the embedding space, similar to \cite{toker2021coming,regmi2019bridging}. In particular, in \cite{toker2021coming}, a ground-view ODI is synthesized from the polar transformation of the satellite view via a GAN, supervised by the corresponding ground-view ground truth. Meanwhile, an extra retrieval branch is applied to constrain the latent representations of two domains. Using conditional GANs, Regmi \etal~\cite{regmi2019bridging} skillfully synthesized the satellite-view image from the ground-view ODI. To learn a robust satellite query representation, they fused the features from the satellite-view synthesis and ground-view ODI, and then matched the query feature with satellite-view features in the embedding space. As the latest work, TransGeo~\cite{zhu2022transgeo} is the first ViT-based framework to extract the position information from the satellite images and ground-view ODIs. With an attention mechanism, TransGeo removes uninformative patches in the satellite-view images and surpasses previous CNN-based methods.

\noindent \textbf{Discussion:} Most cross-view synthesis and geo-localization methods assume that a reference image is precisely centered at the location of any query image. Nonetheless, in practice, the two views are usually not perfectly aligned in terms of orientation~\cite{shi2020looking} and spatial location\cite{zhu2021vigor}. Therefore, how to apply cross-view synthesis and geo-localization methods under challenging conditions is a valuable research direction.

\vspace{-3pt}
\subsubsection{Compression}
\label{sec3.1.3}
Compared with conventional perspective images, omnidirectional data records richer geometrical information with a higher resolution and wider FoV, making it more challenging to achieve effective compression. The early approaches for ODI compression directly utilize the existing perspective methods to compress the perspective projections of the ODIs. For instance, Simone \etal~\cite{Simone2016GeometrydrivenQF} proposed an adaptive quantization method to solve the frequency shift in the viewport image blocks when projecting the ODI to the ERP. By contrast, OmniJPEG~\cite{Rerbek2016JPEGBC} first estimates the region of interest in the ODI and then encodes the ODI based on the geometrical transformation of the region content with a novel format called OmniJPEG, which is an extension of JPEG format~\cite{wallace1992jpeg} and can be viewable on legacy JPEG decoders. Considering the ERP distortion, a graph-based coder is proposed by~\cite{Rizkallah2018RateDO} to adapt the sphere surface. To make the coding progress computationally feasible, the graph partitioning algorithm based on rate distortion optimization~\cite{sullivan1998rate} is introduced to achieve a trade-off between the distortion of reconstructed signals, the signal smoothness on each sub-graph, and the coding cost of partitioning description. 
As a representative CNN-based ODI compression work, OSLO~\cite{bidgoli2021oslo} applies HEALPix~\cite{Gorski2005HEALPixAF} to define a convolution operation directly on the sphere and adapt the standard CNN techniques to the spherical domain. The proposed on-the-sphere representation outperforms the similar learnable compression models on the ERP.

For ODV compression, Li \etal~\cite{Li2019SphericalDR} proposed a representative work aiming to optimize the ODV encoding progress.
They analyzed the distortion impacts of restoring spherical domain signals from the different planar projection types and then applied the rate distortion optimization based on the distortion of signal in spherical domain. Similarly, Wang \etal~\cite{Wang2019SphericalCT} proposed a spherical
coordinates transform-based motion model to address the distortion problem in projections. Another representative method ~\cite{Fu2009TheRD} maps the ODV to the rhombic dodecahedron (RD) map and directly applies the planar perspective videos encoding methods on the RD map. Specifically, the rate control-based algorithms are proposed to achieve better qualities and smaller bitrate errors for ODV compression~\cite{Li2020lambdaDomainPR},~\cite{Zhao2021GameTR}. Zhao \etal~\cite{Zhao2021GameTR} utilized game theory to find optimal inter/intra-frame bitrate allocations while Li \etal~\cite{Li2020lambdaDomainPR} proposed a novel bit allocation algorithm for ERP with the coding tree unit (CTU) level. Similar to~\cite{Su2017LearningSC}, the CTUs in the same row have the same weight to reduce the distortion influence. 

\noindent \textbf{Potential and Challenges:} Based on the aforementioned analysis,  only a few DL-based methods exist in this research domain. Most works combine the traditional planar coding methods with geometric information in the spherical domain. There remain some challenges for DL-based ODI/ODV compression. DL-based image compression methods require the effective metrics as the constraint, \eg, peak signal-to-noise ratio (PSNR), and structural similarity (SSIM). However, due to spherical imaging, traditional metrics are weak to measure the qualities of ODI. Furthermore, the planar projections of the ODI are high memory and distorted, which increase the computation cost and compression difficulty. Future research might consider extending more effective metrics based on the spherical geometric information and restoring a high-quality compressed ODI from a partial input.

\vspace{-4pt}
\subsubsection{Lighting Estimation}
\label{sec3.1.4}
\noindent\textbf{Insight:}
\textit{It aims to predict the high dynamic range (HDR) illumination from low dynamic range (LDR) ODIs.}
%

Illumination recovery is widely employed in many real-world tasks ranging from scene understanding, reconstruction to editing. Hold-Geoffroy \etal~\cite{hold2017deep} proposed a representative framework for outdoor illumination estimation. They first trained a CNN model to predict the sky parameters from viewports of outdoor ODIs, \eg, sun position and atmospheric conditions. They then reconstructed illumination environment maps for the given test images according to the predicted illumination parameters.
Similarly, in~\cite{Gardner2017LearningTP}, a CNN model is leveraged to predict the location of lights in the viewports, and the CNN is fine-tuned to predict the light intensities, \ie, environment maps, from the ODIs.  In~\cite{Gardner2019DeepPI}, geometric and photometric parameters of indoor lighting are regressed from the viewports of ODI, and the intermediate latent vectors are used to reconstruct the environment maps. Another representative method, called EMLight~\cite{zhan2020emlight}, consists of a regression network and a neural projector. The regression network outputs the light parameters, and the neural projector converts the light parameters into the illumination map. In particular, the ground truths of the light parameters are decomposed by a Gaussian map generated from the illumination via a spherical Gaussian function.    

\noindent\textbf{Discussion and Potential:}
From the aforementioned analysis, previous works for lighting estimation on ODIs take a single viewport as the input. The reason might be that the viewports are distortion-less and low-cost with low resolution. However, they suffer from severe drop of spatial information. Hence, it could be beneficial to apply contrastive learning to learn the robust representations from the multiple viewports or components of the tangent images. 

\vspace{-4pt}
\subsubsection{ODI Super-Resolution (SR)}
\label{sec3.1.5}

Existing Head-Mounted Display (HMD) devices~\cite{rolland2005head} require at least the ODI with 21600$\times$10800 pixels for immersive experience, which can not be directly captured by current camera systems~\cite{ozcinar2019super}. One alternative way is to capture low resolution (LR) ODIs and super-resolve them into high resolution (HR) ODIs efficiently. LAU-Net~\cite{Deng2021LAUNetLA}, as the first work to consider the latitude difference for ODI SR, introduces a multi-level latitude adaptive network. It splits an ODI into different latitude bands and hierarchically upscales these bands with different adaptive factors, which are learned via a reinforcement learning scheme. Beyond considering SR on the ERP, Yoon \etal~\cite{yoon2021spheresr} proposed a representative work, SphereSR, to learn a unified continuous spherical local implicit image function and generate an arbitrary projection with arbitrary resolution according to the spherical coordinate queries. For ODV SR, SMFN~\cite{liu2020single} is the first DNN-based framework, including a single-frame and multi-frame joint network and a dual network. The single-frame and multi-frame joint network fuses the features from adjacent frames, and the dual network constrains the solution space to find a better answer.

\vspace{-3pt}
\subsubsection{Upright Adjustment}
\label{sec3.1.6}
\noindent \textbf{Insight:}
\textit{Upright adjustment aims to correct the misalignment of the orientations between the camera and scene to improve the visual quality of ODI and ODV while they are used with a narrow field-of-view (NFoV) display, such as the VR application.}

The standard approach of upright adjustment follows two steps: (i) estimating the position of the pole of the ODI; (ii) applying a rotation matrix to align the estimated north pole. The early representative work~\cite{MichaelBosse2002VanishingPA} estimates the camera rotation according to the geometric structures in the panoramas, \eg, curving straight lines and vanishing points. However, 
these methods are limited to the Manhattan~\cite{vanegas2010building} or Atlanta world~\cite{Schindler2004AtlantaWA} assumption and rely on necessary prior knowledge of geometric structures. Recently, DL-based upright adjustment has been widely studied. Without any specific assumption on the scene structure, DeepUA~\cite{JunhoJeon2018DeepUA} proposes a representative CNN-based framework to estimate the 2D rotations of multiple NFoV images sampled from the ODI and then estimate the 3D camera rotation through the geometric relationship between 3D and 2D rotations. By contrast, Deep360Up~\cite{Jung2019Deep360UpAD} directly takes ERP image as the input and synthesizes the upright version according to the estimated up-vector orientation. In particular, Jung \etal~\cite{Jung2020UprightAW} proposed a two-stage pipeline for ODI upright adjustment. First, the feature map is extracted by a CNN model from the rotated ERP image. The feature map is then mapped into a spherical graph. Finally, a GCN is applied to estimate the 3D camera rotation, which is the location of the point on the spherical surface corresponding to the north pole.

\begin{figure}[t]
    \centering
    \includegraphics[width=1\linewidth]{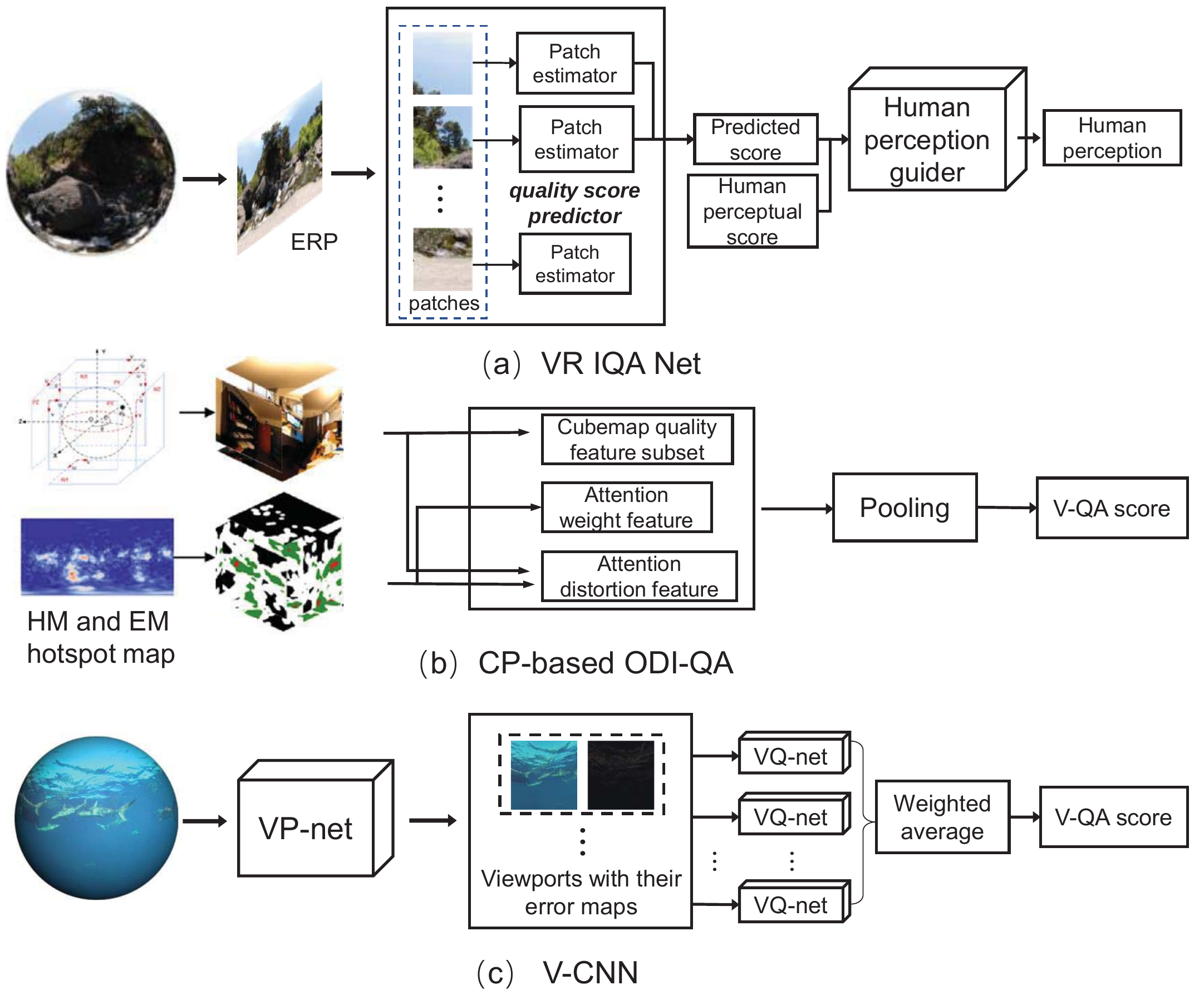}
    \caption{Representative networks for ODI-QA and ODV-QA. (a), (b) and (c) are originally shown in~\cite{Lim2018VRIN},~\cite{Jiang2021CubemapBasedPB} and~\cite{Li2019ViewportPC}.}
    \label{fig:V-CNN}
    \vspace{-10pt}
\end{figure}

\vspace{-5pt}
\subsubsection{Visual Quality Assessment}
\label{sec3.1.7}
Due to the ultra-high resolution and sphere representation of omnidirectional data, visual quality assessment (V-QA) is valuable for the optimization of exiting image/video processing algorithms. We next introduce some representative works on ODI-QA and ODV-QA, respectively.

For the ODI-QA, according to the availability of the reference images, it can be further classified into two categories: full-reference (FR) ODI-QA  and no-reference (NR) ODI-QA. In exiting methods on FR ODI-QA, some works focus on extending the conventional FR image quality assessment metrics, \eg, PSNR and SSIM, to the omnidirectional domain, \eg,~\cite{YuleSun2017WeightedtoSphericallyUniformQE},~\cite{MaiXu2019AssessingVQ}. These works introduce special geometric structures of the ODI and its projection representations to traditional quality assessment metrics and measure the objective quality more accurately. In addition, there are a few DL-based approaches for FR ODI-QA. As the representative work shown in Fig.~\ref{fig:V-CNN}(a), Lim \etal~\cite{Lim2018VRIN,Kim2020DeepVR} proposed a novel adversarial learning framework, consisting of a quality score predictor and a human perception guider, to automatically assess the image quality following the human perception. NR ODI-QA, also called blind ODI-QA, predicts the ODI quality without expensive reference ODIs. 
Considering multi-viewport images in the ERP format, Xu \etal~\cite{Xu2021BlindOI} applied a novel viewport-oriented GCN to process the distortion-less viewports in ERP images and aggregated these features to estimate the quality score via an image quality regressor. A similar strategy is applied in~\cite{Zhou2022NoReferenceQA,Sun2020MC360IQAAM}. By contrast,~\cite{Jiang2021CubemapBasedPB} extracted the features from CP images and their corresponding eye movement (EM) and head movement (HM) hotspot maps and provided a good projection-based potential, that is extracting the features from the multiple projection formats and fusing the features to improve the performance on blind ODI-QA, as shown in Fig.~\ref{fig:V-CNN}(b).

For the ODV-QA, Li \etal~\cite{Li2019ViewportPC} proposed a representative viewport-based CNN approach, including a viewport proposal network and a viewport quality network, as shown in Fig.~\ref{fig:V-CNN}(c). The viewport proposal network generates several potential viewports and their error maps, and viewport quality network rates the V-QA score for each proposed viewport. The final V-QA score is calculated by the weighted average of all viewport V-QA scores.~\cite{Azevedo2020AVM} is another representative that considers the temporal changes of spatial distortions in ODVs and fuses a set of spatio-temporal objective quality metrics from multiple viewports to learn a subjective quality score. Similarly, Gao \etal~\cite{Gao2022QualityAF} modeled the spatial-temporal distortions of ODVs and proposed a novel FR objective metric by integrating
three existing ODI-QA objective metrics. 

\begin{figure}[t!]
    \centering
    \includegraphics[width=0.85\linewidth]{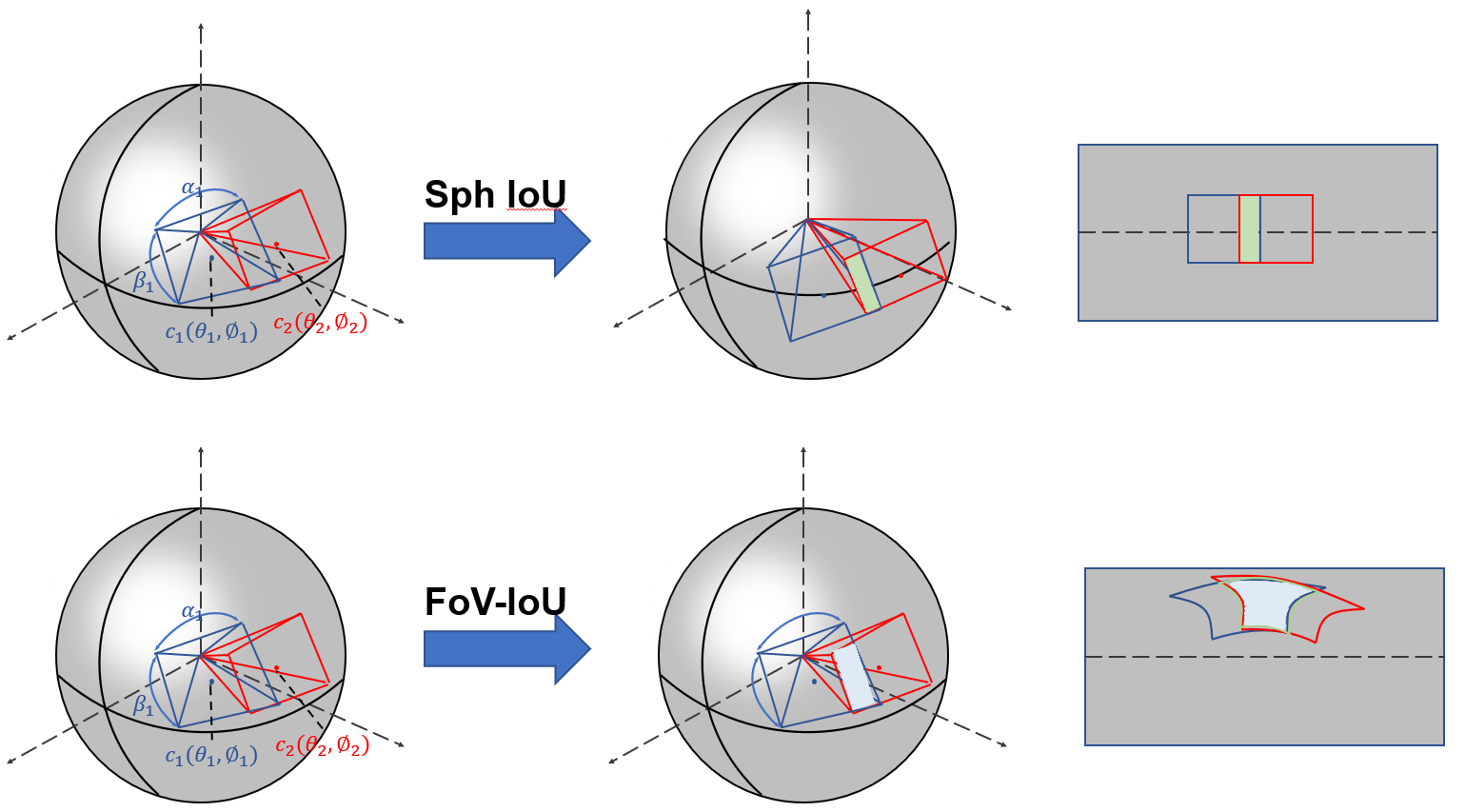}
    \caption{An illustration of Sph IoU~\cite{zhao2020spherical} and FoV-IoU~\cite{Cao2022FieldofViewIF}.}
    \label{fig:iou}
    \vspace{-5pt}
\end{figure}

\vspace{-7pt}
\subsection{Scene Understanding}
\label{sec3.2}
\subsubsection{Object Detection }
\label{sec3.2.1}
Compared with the perspective images, DL-based object detection on ODIs remains two main difficulties: (i) traditional convolutional kernels are weak to process the irregular planar grid structures in the ODI projections; (ii) the criterias adopted in conventional 2D object detection do not fit well to the spherical images. To address the first difficulty, distortion-aware structures are proposed, \eg, multi-scale feature pyramid network in~\cite{Tong2019ObjectDF}, multi-kernel layers in~\cite{wang2019object}. However, the detection flows of these two methods are similar to the methods for 2D domain, which take the whole ERP image as input and predict the regions of interest (ROIs) to obtain the final bounding boxes. Considering the wide FoV of ERP, Yang \etal~\cite{yang2018object} proposed a representative framework, which can leverage the conventional 2D images to train a panoramic detector. The detecting progress consists of three sub-steps: stereo-projection, YOLO detectors, and bounding box post processing. Especially, they generated four stereographic projections with a $180^\circ \times 180^\circ$ FoV from an ERP and the four result maps predicted by the YOLO detectors. Finally, the sub-window detected bounding boxes are re-projected to the ERP and re-aligned into the final distortion-less bounding boxes. 

To tackle the second difficulty, a novel kind of spherical bounding boxe (SphBB) and spherical Intersection over Union (SphIoU) for ODI object detection are introduced in~\cite{zhao2020spherical}, as shown in the first row of Fig.~\ref{fig:iou}. SphBB is represented by the coordinates $\theta$, $\phi$ of the object centers and the unbiased FoVs $\alpha$, $\beta$ of the objective occupation. SphIoU is similar to planar IoU and calculated by the IoU between two SphBBs. Concretely, FoVBBs are moved to the equator that is undistorted. Similarly, Cao \etal~\cite{Cao2022FieldofViewIF} proposed a novel IoU calculation method without any extra movement, called FoV-IoU. As shown in the second row of Fig.~\ref{fig:iou}, FoV-IoU better approximates the exact computation of IoU between two FoV-BBs compared with the SphIoU.

\begin{figure}[t!]
    \centering
    \includegraphics[width=.98\linewidth]{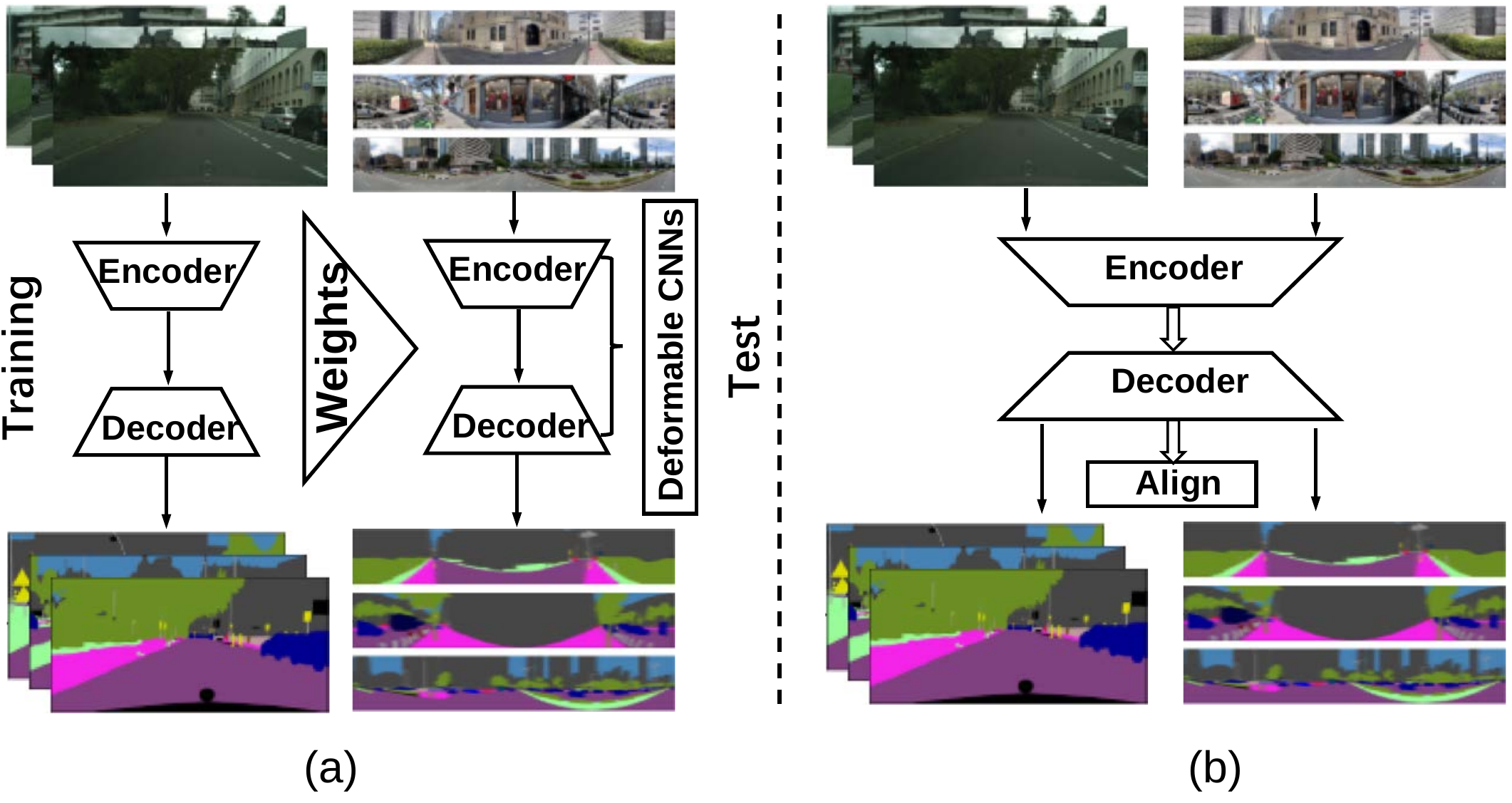}
    \caption{Representative methods for unsupervised ODI semantic segmentation. (a) Unsupervised learning with deformable CNNs \cite{Tateno2018DistortionAwareCF}. (b) Unsupervised learning with domain adaptation~\cite{MaZYRS21}.}
    \vspace{-5pt}
\end{figure}

\vspace{-3pt}
\subsubsection{Semantic Segmentation}
\label{sec3.2.2}
\begin{table*}[htbp]
\setlength{\abovecaptionskip}{-0.1cm}
\caption{Semantic segmentation by some representative methods. “S”: supervised, "U": Unsupervised, “D”: domain adaptation.}
\begin{center}
\renewcommand\arraystretch{1.5}
   \begin{tabularx}{0.97\textwidth}{c|c|c|c|c|c|X<{\centering}}
    \hline
    Method  & Publication & Input& Dataset   & Deformable & Supervision  & Highlight \\
    \hline
    \hline
    Tateno~\cite{Tateno2018DistortionAwareCF} &ECCV'2018& ERP& Stanford2d3d&$\checkmark$&U&Distortion-aware convolution\\
    Zhang~\cite{zhang2019orientation} & ICCV'2019& Tangent&  Stanford2D3D/Omni-SYNTHIA& $\checkmark$ &S& Orientation-aware convolutions\\
    Lee~\cite{Lee2019SpherePHDAC} & CVPR'2019 & Tangent& SYNTHIA/Stanford2D3D & $\checkmark$ & S& Icosahedral geodesic polyhedron\\
    Viu~\cite{Guerrero-ViuFDG20} &ICRA'2020& ERP&SUN360& $\checkmark$&S& Equirectangular convolutions\\
    Yang~\cite{Yang2021CapturingOC}&CVPR'2021& ERP& PASS/WildPASS& \ding{55}&U&Concurrent attention networks\\
    Zhang~\cite{Zhang2022BendingRD}& CVPR'2022& ERP&Stanford2D3D/DensePASS&$\checkmark$ &U & Deformable MLP\\
    Zhang~\cite{zhang2021transfer} & T-ITS'2022 &ERP& DensePASS/VISTAS&\ding{55}&D& Uncertainty-aware adaptation          \\
    
    \hline
\end{tabularx}
\end{center}
\label{table:segmentation}
\vspace{-0.3cm}
\end{table*}

DL-based omnidirectional semantic segmentation has been widely studied because ODI can encompass exhaustive information about the surrounding space. There are many practically remaining challenges, \eg, distortions in the planar projections, object deformations, computed complexity, and scarce labeled data. We next introduce some representative methods for ODI semantic segmentation via supervised learning and unsupervised learning.

Due to the lack of real-world datasets, Deng \etal~\cite{Deng2017CNNBS} firstly generated ODIs from an existing dataset of urban traffic scenes and then designed an approach, called zoom augmentation, to transform the conventional images into fisheye images. Meanwhile, they proposed a CNN-based framework with a special pooling module to integrate the local and global context information and handle complex scenes in the ODIs. Considering that CNNs have inherently limited ability to handle the distortions in ODIs, Deng \etal~\cite{DengYLLHW20} proposed a method, called Restricted Deformable Convolution, to model geometric transformations and learn a convolutional filter size from the input feature map. Zoom augmentation was also applied to~\cite{DengYLLHW20} for enriching the train data. As the first framework to conduct semantic segmentation on the real-world outdoor ODIs, SemanticSO~\cite{Orhan2022SemanticSO} builds a distortion-aware CNN model using the equirectangular convolutions~\cite{FernandezLabrador2020CornersFL}. 

Due to the time-consuming and expensive cost of ground truth annotations for ODIs, endeavours have been made to synthesize ODI datasets from the conventional images and utilize knowledge transfer to adopt models directly trained with the perspective images. PASS~\cite{Yang2019CanWP} is the first work to bypass fully dense panoramic annotations and aggregate features represented by conventional perspective images to fulfill the pixel-wise segmentation in panoramic imagery. Based on the PASS, DS-PASS~\cite{0001HCXWS20} further re-uses the knowledge learned from perspective images and adapts the model learned from the 2D domain to panoramic domain. Meanwhile, in DS-PASS, the sensitivity to spatial details is enhanced by implementing attention-based lateral connections to perform segmentation accurately. To reduce the domain gap between the ODI and perspective image, Yang \etal~\cite{Yang2021CapturingOC} proposed a representative cross-domain transfer framework that designs an efficient concurrent attention network to capture the long-range dependencies in ODI imagery and integrates the unlabeled ODIs and labeled perspective images into training. A similar strategy was applied in~\cite{JausYS21},~\cite{YangHFWS22} and~\cite{MaZYRS21}. Particularly, in~\cite{MaZYRS21}, a shared attention module is used to extract features from the 2D domain and panoramic domain, and two domain adaption modules are used to "teach" the panoramic branch by the perspective branch. For unsupervised semantic segmentation, there also exist some works considering the geometric structure of ODI~\cite{Tateno2018DistortionAwareCF}. For instance, Zhang \etal~\cite{zhang2019orientation} proposed an orientation-aware CNN framework based on the icosahedron mesh representation of ODI and introduced an efficient interpolation approach of the north-aligned kernel convolutions for features on the sphere.

\vspace{-3pt}
\subsubsection{Monocular Depth Estimation }
\label{sec3.2.3}

\begin{table*}[h]
\setlength{\abovecaptionskip}{-0.1cm}
\caption{Monocular depth estimation by some representative methods. “S”: supervised, “D”: domain adaptation.}
\begin{center}
\renewcommand\arraystretch{1.5}
    \begin{tabularx}{0.97\textwidth}{c|c|c|c|c|X<{\centering}}
    \hline
    Method & Publication & Supervision & Input types & Architecture & Loss functions \\
    \hline
    \hline
    Zioulis~\cite{zioulis2018OmniDepthDD} & ECCV'18 & S & ERP & Rectangular filters & l2 loss+smooth loss \\
    Pintore~\cite{Pintore2021SliceNetDD} & CVPR'21 & S & ERP & Slice-based representation and LSTM & BerHu loss~\cite{laina2016deeper} \\
    Zhuang~\cite{zhuang2021acdnet} & AAAI'22 & S & ERP & Dilated filters & BerHu loss \\
    Wang~\cite{wang2020bifuse} & CVPR'20 & S & ERP+CP & Two-branch network and bi-projection fusion & BerHu loss \\
    Rey-Area~\cite{reyarea2021360monodepth} & CVPR'22 & S & Tangent & Perspective network+Alignment+Blending & Energy function \\
    Li~\cite{li2022omnifusion} & CVPR'22 & S & Tangent & Geometric embedding+Transformer & BerHu loss \\
    Jin~\cite{jin2020geometric} & CVPR'20 & S & ERP & Structure information as prior and regularizer & l1 loss+cross entropy loss \\
    \hline
    Wang~\cite{wang2018self} & ACCV'18 & Self-S & CP & Depth estimation+camera motion estimation & photometric + pose loss \\
    Zioulis~\cite{zioulis2019spherical}  & 3DV'19 & Self-S & ERP & View synthesis in horizontal, vertical and trinocular ones & photometric +smooth loss \\
    Yun~\cite{yun2021improving} & AAAI'22 & S+Self-S & ERP & ViT+pose estimation    & \makecell[c]{SSIM~\cite{wang2004image}+gradient \\ +L1+photometric loss} \\
    \hline
    Tateno~\cite{Tateno2018DistortionAwareCF}  & ECCV'18 & D & ERP & Distortion-aware filters & BerHu loss \\
\hline
\end{tabularx}
\end{center}
\label{table:depth}
\vspace{-0.3cm}
\end{table*}
\begin{figure*}[htbp]
    \centering
    \includegraphics[width=1\textwidth]{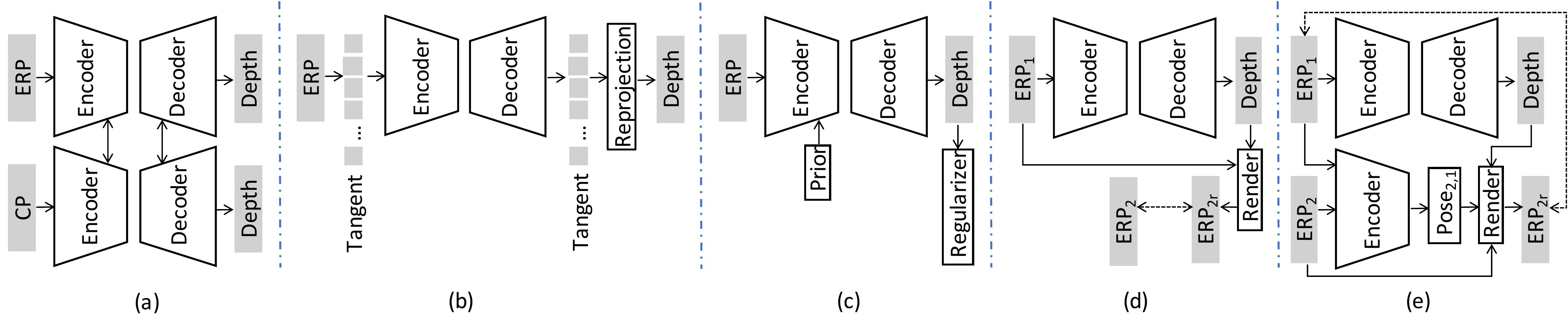}
    \caption{Representative monocular depth estimation methods. (a) Dual-branch methods with ERP and CP as inputs. (b) Methods taking tangent images as input and re-project them into ERP images. (c) Methods utilizing extra geometric information as the prior and regularizer. (d) View synthesis methods with certain baselines. (e) Self-supervised multi-frame methods with the pose estimation.} 
    \label{fig:Monocular depth estimation}
    \vspace{-0.4cm}
\end{figure*}
Thanks to the emergence of large-scale panoramic depth datasets, monocular depth estimation has evolved rapidly. As shown in Fig.~\ref{fig:Monocular depth estimation}, there are several trends: (i) Tailored networks, \eg, distortion-aware convolution filters~\cite{Tateno2018DistortionAwareCF} and robust representations~\cite{Pintore2021SliceNetDD}; (ii) Different projection types of ODIs~\cite{wang2020bifuse},~\cite{feng2022360},~\cite{li2022omnifusion}, as depicted in Fig.~\ref{fig:Monocular depth estimation}(a), (b); (iii) Inherent geometric priors~\cite{eder2019pano},~\cite{jin2020geometric}, as shown in Fig.~\ref{fig:Monocular depth estimation}(c); (iv) Multiple views~\cite{zioulis2019spherical} or pose estimation~\cite{yun2021improving}, as shown in Fig.~\ref{fig:Monocular depth estimation} (d), (e), respectively.

\noindent \textbf{\textit{Tailored networks:}} To reduce the influence of the stretch distortion, Zioulis \etal~\cite{zioulis2018OmniDepthDD} proposed the first work by directly using the ODIs. It follows \cite{Su2017LearningSC} to transfer regular square convolution filters into row-wise rectangles and vary filter sizes to address the distortions at the poles. Tateno \etal~\cite{Tateno2018DistortionAwareCF} proposed a deformable convolution filter that samples the pixel grids on the tangent planes according to unit sphere coordinates. Recently, Zhuang \etal~\cite{zhuang2021acdnet} proposed a novel framework to combine different dilated convolutions and extend the receptive field in the ERP images. In comparison, Pintore \etal~\cite{Pintore2021SliceNetDD} proposed a framework, named SliceNet, with regular convolution filters to work on the ERP directly. SliceNet reduces the input tensor only along the vertical direction to collect a sequence of vertical slices and adopts an LSTM~\cite{shi2015convolutional} network to recover the long- and short-term spatial relationships among slices.

\noindent \textbf{\textit{Different projection formats:}} There are some attempts to address the distortion in the ERP via other distortion-less projection formats, \eg, CP, tangent projection. As a representative work, BiFuse~\cite{wang2020bifuse} introduces a two-branch pipeline, where one branch processes the ERP input and another branch extracts the features from CP, to simulate the peripheral and foveal vision of human, as shown in Fig.~\ref{fig:Monocular depth estimation} (a). Then, a fusion model is proposed to combine the semantic and geometric information of the two branches. Inspired by BiFuse, UniFuse~\cite{jiang2021unifuse} designs a more effective fusion module to combine the two kinds of features and unidirectionally feeds the CP features to the ERP features only at the decoding stage. To better extract the global context information, GLPanoDepth~\cite{bai2022glpanodepth} converts ERP input into a set of CP images and then exploits a ViT model  to learn the long-range dependencies. As the tangent projection produces less distortion than CP, 360MonoDepth~\cite{reyarea2021360monodepth} trains the SoTA depth estimation models in 2D domain~\cite{ranftl2021vision} with tangent images and re-projects predicted tangent depth maps into the ERP with alignment and blending, as shown in Fig.~\ref{fig:Monocular depth estimation}(b). However, directly re-projecting the tangent images back to the ERP format will cause overlapping and discontinuity. Therefore, OmniFusion~\cite{li2022omnifusion} (the SoTA method by far) introduces additional 3D geometric embeddings to mitigate the discrepancy in patch-wise features and aggregates patch-wise information with an attention-based transformer.

\noindent \textbf{\textit{Geometric Information Prior:}} Some methods add extra geometric information priors to improve the performance, \eg, edge-plane information, surface normal, boundaries, as shown in Fig.~\ref{fig:Monocular depth estimation}(c). Eder \etal~\cite{eder2019pano} assumed that each scene is piecewise planar and the principal curvature of each planar region, which is the second derivative of depth, should be zero. Consequently, they proposed a plane-aware learning scheme that jointly predicts depth, surface normal, and boundaries. Similar to~\cite{eder2019pano}, Feng \etal~\cite{feng2020deep} proposed a framework to refine depth estimation using the surface normal and uncertainty scores. For a pixel with higher uncertainty, its prediction is mainly aggregated from the neighboring pixels. Particularly, Jin \etal~\cite{jin2020geometric} demonstrated that the representations of geometric structure, \eg, corners, boundaries, and planes, can provide the regularization for depth estimation and benefit it as the prior information well.

\noindent \textbf{\textit{Multiple Views:}} As ODI depth annotations are expensive, some works leverage the multiple viewpoints to synthesize data and obtain competitive results. Zioulis \etal~\cite{zioulis2019spherical} explored the spherical view synthesis for self-supervised monocular depth estimation. As shown in Fig.~\ref{fig:Monocular depth estimation}(d), in~\cite{zioulis2019spherical}, after predicting the ERP format depth map, stereo viewpoints in vertical and horizontal baselines are synthesized by the depth-image-based rendering. Synthesized images are supervised by real images with the same viewpoints via photometric image reconstruction loss. To improve accuracy and stability simultaneously, Yun \etal~\cite{yun2021improving} proposed a joint learning framework to estimate monocular depth via supervised learning and estimate poses via self-supervised learning from the adjacent frames of ODV, as shown in Fig.~\ref{fig:Monocular depth estimation}(e).

\noindent \textbf{Discussion:} Based on the aforementioned analysis, most methods only consider indoor scenes due to two main reasons: (i) Some geometric priors are ineffective in the wild, \eg, the plane assumption; (ii) Outdoor scenes are more challenging due to the scale ambiguity in approximately infinite regions (\eg, sky), and objects in various shapes and sizes~\cite{feng2022360}.

It has been demonstrated that directly applying the DL-based methods for 2D optical flow estimation on ODI will obtain the unsatisfactory results~\cite{Apitzsch2018Cubes3DNN}.  To this end, Xie \etal~\cite{Xie2019EffectiveCN} introduced a small diagnostic dataset FlowCLEVR and evaluated the performance of three kinds of tailored convolution filters, namely the correlation, coordinate and deformable convolutions, for estimating the omnidirectional optical flow. The domain adaptation frameworks~\cite{Artizzu2021OmniFlowNetAP, Bhandari2021RevisitingOF} benefit from the development of optical flow estimation in the perspective domain. Similar to~\cite{Apitzsch2018Cubes3DNN}, OmniFlowNet~\cite{Artizzu2021OmniFlowNetAP} is built on FlowNet2 and the convolution operation is inspired by~\cite{FernandezLabrador2020CornersFL}. Especially, as the extension of~\cite{Hui2018LiteFlowNetAL}, LiteFlowNet360~\cite{Bhandari2021RevisitingOF} uses kernel transformation techniques to solve the inherent distortion problem caused by the sphere-to-plane projection. A representative pipeline is proposed by~\cite{Shi2022PanoFlowLO}, consisting of a data augmentation method and a flow estimation module. The data augmentation method overcomes the distortions introduced by ERP, and the flow estimation module exploits the cyclicity of spherical boundaries to convert long-distance estimation into a relatively short-distance estimation.

\vspace{-5pt}
\subsubsection{Video Summarization}
\label{sec3.2.5}
\noindent\textbf{Insight:} \textit{Video summarization aims to generate representative and complete synopsis by selecting the parts containing the most critical information of the ODV.}

Compared with the methods for 2D video summarization, only a few works have been proposed for ODV summarization. Pano2Vid~\cite{SuJG16} is the representative framework that contains two sub-steps: detecting candidate events of interest in the entire ODV frames and applying dynamic programming to link detected events. However, Pano2Vid requires observing the whole video and is less capable for video streaming applications. Deep360Pilot~\cite{hu2017deep} is the first framework to design a human-like online agent for automatic ODV navigation of viewers. Deep360pilot consists of three steps: object detection to obtain the candidate objects of interest, training RNN to choose the important object, and capturing exciting moments in ODV. AutoCam~\cite{SuG17} generates the normal NFoV videos from the ODVs following human behavior understanding. An similar strategy was applied by Yu \etal~\cite{YuLNKK18}. They built a deep ranking model for spatial summarization to select NFOV shots from each frame in the ODV and generated a spatio-temporal highlight video by extending the same model to the temporal domain. Moreover, Lee~\cite{LeeSYK18} proposed a novel deep ranking neural network model for summarizing ODV both spatially and temporally. 

\noindent \textbf{Discussion:} Based on the above analysis, only a few methods exist in this research domain. As a temporal-related task, applying the transformer mechanism to ODV summarization could be beneficial. In addition, previous works only considered the ERP format, which suffer from the most severe distortion problems. Therefore, it is better to consider the CP, tangent projection or sphere format as the input for ODV summarization.

\vspace{-5pt}
\subsection{3D Vision}
\label{sec3.3}
\begin{table*}[hbtp]
\setlength{\abovecaptionskip}{-0.1cm}
\caption{Room Layout estimation overview on representative studies.}
\begin{center}
\renewcommand\arraystretch{1.7}
    \begin{tabularx}{0.92\textwidth}{c|c|c|c|c|c}
    \hline
    Method & Publication & Architecture & Highlight & Projection & Task \\
    \hline
    \hline
Zhang \cite{Zhang2021DeepPanoContextP3} & ICCV’21 & \makecell[c]{Mask RCNN+ODN\\+LIEN+HorizonNet}& Context relation modeling & ERP &\makecell[c]{ Layout+ object\\+semantic labels} \\
Yang \cite{Yang2019DuLaNetAD} & CVPR’19 & Two ResNet on ceiling and floor & Projection feature fusion & \makecell[c]{ ERP,\\ceiling} & Layout \\
Zou \cite{Zou2018LayoutNetRT} & CVPR’18 & CNN+3D layout regressor & Boundary+Corner map prediction & ERP & Layout \\
Tran \cite{Tran2021SSLayout360SI} & CVPR’21 & HorizonNet+EMA & Semi-supervised learning & ERP & Layout \\
Pintore \cite{Pintore2020AtlantaNetIT} & ECCV’20 & ResNet+RNN & Atlanta World indoor Model & \makecell[c]{ ERP, \\ceiling}  & Layout~ \\
Sun \cite{Sun2019HorizonNetLR} & CVPR’19 & ResNet+RNN & 1D representation of layout & ERP & Layout \\
Sun \cite{Sun2021HoHoNet3I} & CVPR’21 & \makecell[c]{ResNet+\\efficient height compression} & Latent horizontal feature & ERP &\makecell[c]{ Layout, depth\\+semantic labels} \\
Wang \cite{Wang2021LED2NetM3} & CVPR’21 & HorizonNet$\&$L2D transformation & Differentiable depth rendering & ERP & Layout \\
\hline
\end{tabularx}
\end{center}
\label{table:room layout}
\vspace{-0.5cm}
\end{table*}
\subsubsection{Room Layout estimation and Reconstruction}

\label{sec3.3.1}

\noindent\textbf{Insight:} \textit{Room Layout estimation and reconstruction consists of multiple sub-tasks such as layout estimation, 3D object detection and 3D object reconstruction. This comprehensive task aims to facilitate holistic scene understanding based on a single ODI.}

\begin{figure}[t!]
    \centering
    \includegraphics[width=.94\columnwidth]{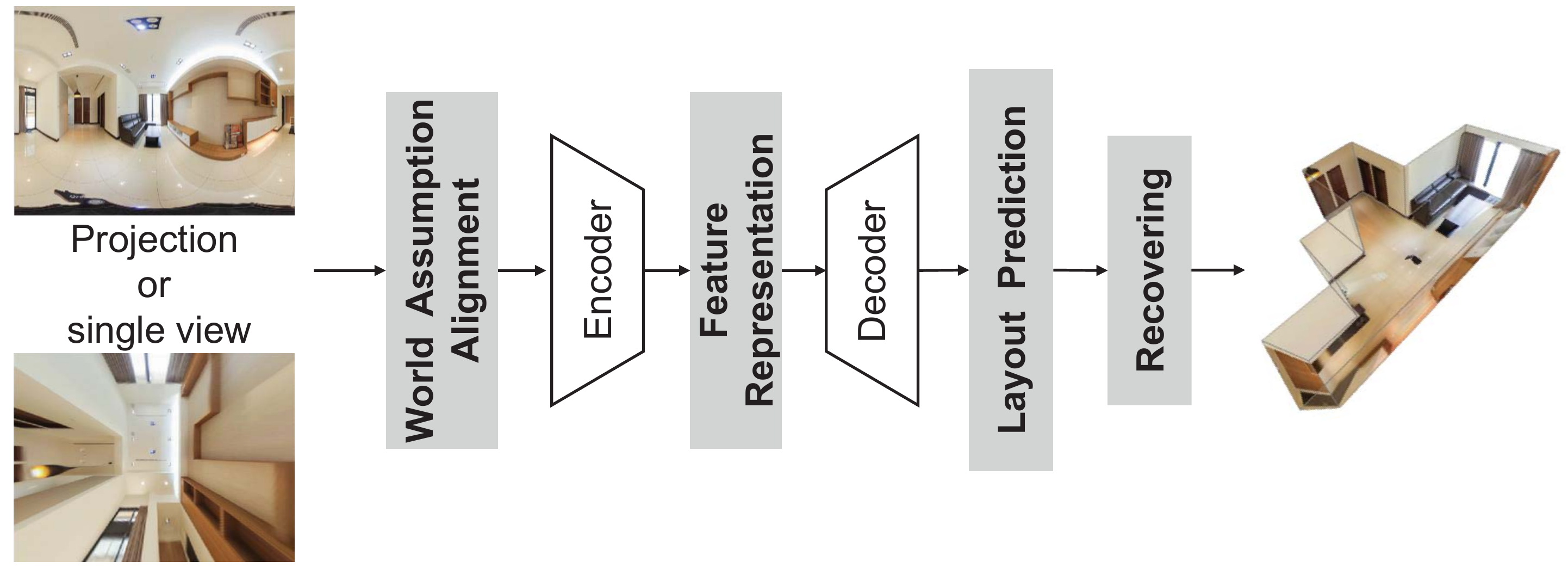}
    \caption{The general room layout prediction architecture.}
    \label{fig:room layout}
    \vspace{-12pt}
\end{figure}

As the indoor panoramas can cover wider surrounding environment and capture more context cues than conventional perspective images, they are beneficial to scene understanding and widely applied into room layout estimation and reconstruction. Zou \etal~\cite{Zou2021ManhattanRL} summarized that the general procedure of layout estimation and reconstruction contains three sub-steps: edge-based alignment, layout elements prediction, and 3D layout elements recovery, as shown in Fig.~\ref{fig:room layout}. The representative work, proposed by Zhang \etal~\cite{Zhang2021DeepPanoContextP3}, conducts the first DL-based pipeline for holistic 3D scene understanding that recovers 3D room layout and detailed information, \eg, shape, pose, and location of objects from a single ODI. In~\cite{Zhang2021DeepPanoContextP3}, a context-based GNN is designed to predict the relationships across the objects and room layout and achieves the SoTA performance on both geometry accuracy of room layout and 3D object arrangement.

For the alignment, this pre-possessing step provides indoor geometric information as the prior knowledge to ease the network training. Several SoTA approaches~\cite{Yang2019DuLaNetAD, Zou2018LayoutNetRT, Tran2021SSLayout360SI} follow the "Manhattan world" assumption, in which all walls are aligned with a canonical coordinate system, and the floor plane direction is estimated by selecting long line segments and voting for the three mutually orthogonal vanishing directions. In contrast, AtlantaNet~\cite{Pintore2020AtlantaNetIT} predicts the 3D layout from less restrictive scenes that are not limited to "Manhattan World" assumption. AtlantaNet follows "Atlanta World" assumption and projects an gravity-aligned ODI into two
horizontal planes to predict a 2D room footprint on the floor plan and a room height to recover the 3D layout.

For the layout element prediction, the primary task is to estimate layout boundaries and corner positions. On the one hand, the related methods usually choose different projections  of ODIs as the input. For instance, 
some methods~\cite{Zou2018LayoutNetRT,Sun2019HorizonNetLR,Sun2021HoHoNet3I} predict the layout only from ERP images. Besides the ERP, Yang \etal~\cite{Yang2019DuLaNetAD} added a perspective ceiling-view image, which is obtained from the ERP through an equirectangular-to-perspective (E2P) conversion, as an extra input. They then extracted the features from the two formats by a two-branch network and fused the two-modal features to predict the layout elements. The advantage of~\cite{Yang2019DuLaNetAD} is that it directly uses the multi-projection model to jointly predict a Manhattan-world floor plan instead of estimating the number of corners. On the other hand, recent methods varied in their ways of feature representation. For instance, HorizonNet~\cite{Sun2019HorizonNetLR} represents the room layout of the ODI as three 1D embedding vectors and recovers 3D room layouts from 1D predictions with low computation cost. Differently, Wang \etal~\cite{Wang2021LED2NetM3} converted the layout into 'horizon-depth' through ray casting of a few points. This transformation maintains the simplicity of layout estimation and improves the generalization capacity to unseen room layouts. 

For final recovery, the general strategy~\cite{Zou2018LayoutNetRT, Yang2019DuLaNetAD, Sun2019HorizonNetLR} is to reconstruct the layout by the optimization of mapping each pixel between walls and corners. In particular, it defines the weighted loss of probability maps of floors, ceilings, and corners. The major difficulty is the layout boundary occlusions when the camera position is not ideal for the entire display. To address this problem, HorizonNet~\cite{Sun2019HorizonNetLR} observes the occlusions by examining the orientation of the first Principal Component Analysis (PCA) component of adjacent walls and recovers occluded parts according to the long-term dependencies of global geometry.

\vspace{-6pt}
\subsubsection{Stereo Matching}
\label{sec3.3.2}

Human binocular disparity depends on the difference between the projections on the retina, that is, a sphere projection rather than a planar projection. Therefore, stereo matching on the ODIs is more similar to the human vision system. In ~\cite{Seuffert2021ASO}, they discussed the influence of omnidirectional distortion on the CNN-based methods and compared the quality of disparity maps predicted from the perspective and omnidirectional stereo images. The experimental results show that stereo matching based on the ODIs is more advantageous for numerous applications, \eg, robotics, AR/VR, and several other applications. General stereo matching algorithms follow four steps: (i) matching cost computation, (ii) cost aggregation, (iii) disparity computation with optimization, and (iv) disparity refinement. As the first DNN-based omnidirectional stereo framework, SweepNet~\cite{Won2019SweepNetWO} proposes a wide-baseline stereo system to compute the matching cost map from a pair of images captured by cameras with ultra-wide FoV lenses and uses a global sphere sweep at the rig coordinate system to generate an omnidirectional depth map directly. By contrast,  OmniMVS~\cite{Won2019OmniMVSEL} takes four 220$^\circ$ FoV fisheye views as the input to train an end-to-end DNN model and uses a 3D encoder-decoder block to regularize the cost volume. The method proposed in~\cite{Won2021EndtoEndLF}, as the extension of OmniMVS, provides a novel regularization of cost volume based on the uncertainty of prior guidance. Another representative work, 360SD-Net~\cite{Wang2020360SDNet3S}, is the first end-to-end trainable network for omnidirectional stereo depth estimation with the top-bottom ODI pairs as the input. It mitigates the distortion in the ERP images through an additional polar angle coordinate input and a learnable cost volume.

\subsubsection{SLAM}
\label{sec3.3.3}

SLAM is an intricate system that adopts multiple cameras, \eg, monocular, stereo, or RGB-D, combined with sensors onboard a mobile agent to reconstruct the environment and estimate the agent pose in real-time. SLAM is often used in real-time navigation and reality augmentation, \eg, Google Earth. The stereo information, such as key points~\cite{Chiuso2002StructureFM} and dense or semi-dense depth maps\cite{MurArtal2017ORBSLAM2AO}, is indispensable to build an accurate modern SLAM system. Specifically, compared with traditional monocular SLAM~\cite{MurArtal2015ORBSLAMAV} or multi-view SLAM~\cite{Urban2016MultiColSLAMA}, the omnidirectional data can provide the richer texture and structure information due to a large FoV, and the omnidirectional SLAM avoids the influence of discontinued frames in the surrounding environment and enjoys the technical advantage of complete positioning and mapping. Caruso \etal~\cite{Caruso2015LargescaleDS} proposed a representative monocular SLAM method for omnidirectional cameras in which the direct image alignment and pixel-wise distance filtering are directly formulated.
Zachary \etal~\cite{ZacharyTeed2021DROIDSLAMDV} proposed a general framework that accepts multiple types of sensor data and is capable of iterative updates of camera pose and pixel-wise depth. DeepFactors~\cite{Czarnowski2020DeepFactorsRP} performs joint optimization of the pose and depth variables to detect the loop closure. As the omnidirectional data has rich geometry and texture information, further works may consider how to cultivate the full potential of DL and utilize these imaging advantages to construct a fast and accurate SLAM system.

\vspace{-7pt}
\subsection{Human Behavior Understanding  }
\label{sec3.4}

\subsubsection{Saliency Prediction }
\label{sec3.4.1}
\begin{table*}[hbtp]
\setlength{\abovecaptionskip}{-0.1cm}
\caption{Deep ODI and ODV saliency prediction by some representative methods. EM and HM mean eye and head movement.}
\begin{center}
\renewcommand\arraystretch{1.7}
    \begin{tabularx}{0.97\textwidth}{c|c|c|c|c|c|c}
    \hline
Method                                                          & Input                      & Publication                                         & EM          & HM         & Highlight                                                           & Contribution                                                                                                                                      \\ \hline
\hline
 Dai~\cite{Dai2020DilatedCN}            &  IMG &  ICASSP'20                  & $\checkmark$  &$\checkmark$   &  CP $\&$ 2D CNN                                &  Dilated convolution                                                            \\
 Lv~\cite{Lv2020SalGCNSP}               &  IMG &  ACM MM'20                  & $\checkmark$  &  $\checkmark$ &  Spherical images $\&$ GCN                     &  GCN with spherical interpolation                                                 \\
 Chao~\cite{Chao2021AMV}                    & IMG & TMM'21                         &   $\checkmark$ &  $\checkmark$  &  Multi-viewports$\&$ 2D CNN     & Different FoV viewports                  \\

Abdelaziz~\cite{Abdelaziz2021Rethinking3I} & IMG & ICCV'21                      & $\checkmark$   &  $\checkmark$  & \makecell[c]{ERP $\&$ 2D CNN \\ $\&$ self-attention mechanism}        & \makecell[c]{Contrastive learning \\ to maximize the mutual information }                       \\
Xu~\cite{MaiXu2021SaliencyPO}                 & IMG & TIP'21                         & \xmark   & $\checkmark$ & ERP $\&$ deep reinforcement learning           & Generative adversarial imitation learning                                                            \\ \hline
Nguyen~\cite{Nguyen2018YourAI}             & VID & ACM MM'18                      & \ding{55} & $\checkmark$ & ERP $\&$ 2D CNN  $\&$  LSTM                            & Transfer learning    \\
Chen~\cite{cheng2018cube}                & VID & CVPR'18                        & \ding{55} & $\checkmark$  & CP $\&$ 2D CNN $\&$ convLSTM                         & Spatial-temporal network $\&$ Cube Padding                   \\
Zhang~\cite{Zhang2018SaliencyDI}           & VID & ECCV'18                        & $\checkmark$ & $\checkmark$ & ERP $\&$ spherical CNN                            & Spherical crown convolution kernel                     \\
Xu~\cite{Xu2019PredictingHM}               & VID & TPAMI'19                       & \ding{55} & $\checkmark$ & ERP $\&$ deep reinforcement learning              & Deep reinforcement learning         \\
Zhu~\cite{Zhu2021ViewingBS}                & VID & TCSVT'21                       & $\checkmark$ &\ding{55}  & Image patches $\&$ GCN                            & Graph convolution and feature alignment                                                         \\
Qiao~\cite{Qiao2021ViewportDependentSP}    & VID & TMM'21                         & $\checkmark$ & \ding{55}& Multi-viewports$\&$ 2D CNN $\&$ convLSTM     & Multi-Task Deep Neural Network                                 \\
\hline
\end{tabularx}
\end{center}
\label{table:saliency}
\vspace{-12pt}
\end{table*}



\begin{figure}[t!]
    \centering
    \includegraphics[width=.92\linewidth]{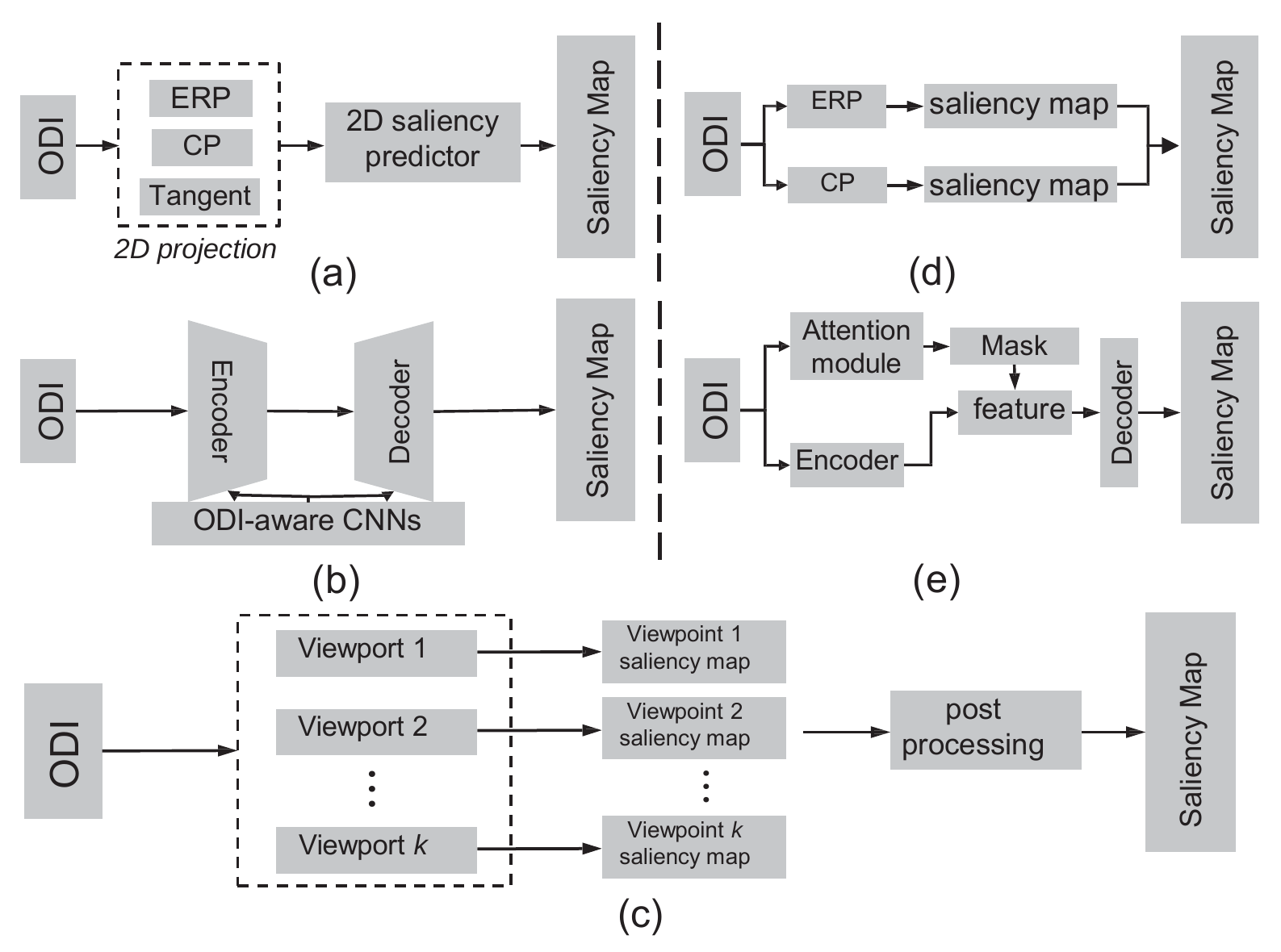}
    \caption{ Deep saliency prediction methods. (a) Directly using traditional 2D models on planar projections of ODI. (b) Using specific ODI-aware CNNs to predict the omnidirectional saliency maps. (c) Aggregating the saliency maps predicted by several viewport images. (d) Combining the predicted saliency maps from different projection types. (e) Using attention mechanism.}
    \label{fig:saliency}
    \vspace{-12pt}
\end{figure}

Recently, there have been several research trends in ODI saliency prediction, building on DL progress: (i) From 2D traditional convolutions to 3D specific convolutions; (ii) From single feature to multiple features; (iii) From single ERP input to multi-type inputs; (iv) From normal CNN-based learning to novel learning strategies.
In Table.~\ref{table:saliency}, numerous DL-based methods have been proposed for ODI saliency prediction. In the following, we introduce and analyze some representative networks, as shown in Fig.~\ref{fig:saliency}.

(i) To directly apply 2D deep saliency predictors on ODIs and reduce the unsatisfactory distortion in ODIs, many works~\cite{ SalNet360, Dai2020DilatedCN} convert ODIs into 2D projection format. As the first attempt of DNNs on ODI saliency prediction, SalNet360~\cite{SalNet360} subdivides an ERP into a set of six CP patches as the input because CP avoids the heavy distortions near the poles like ERP. Then SalNet360 combines predicted saliency maps and per-pixel spherical coordinates of these patches to output a resulting saliency map in ERP format. Differently, a few works~\cite{Zhang2022360degreeVS, Lv2020SalGCNSP} propose the ODI-aware convolution filters for saliency prediction, and learn the relationships between the features from a non-distorted space. The representative work, SalGCN~\cite{Lv2020SalGCNSP}, transfers the ERP image to a spherical graph signal representation, generates the spherical graph signal representation of the saliency map and finally reconstructs the ERP format saliency map through the spherical crown-based interpolation. SalGFCN~\cite{Yang2021SalGFCNGB} proposes a SoTA method that is composed of a residual U-Net architecture based on the dilated graph convolutions and attention mechanism in the bottleneck.
(ii) The viewports are the rectangular windows on ERP with different narrow FoVs caused by observers' head movement. Due to less distortions in viewports, some works~\cite{ Mazumdar2019ACA, Suzuki2018SaliencyME} choose a set of viewports on ERP as the input and extract the multiple independent features from these viewports. The final omnidirectional saliency map is generated by a set of viewport saliency maps and refined via an equator biased post-processing. Different from most prior multi-feature works extracting the low-level geometric features, Mazumdar \etal~\cite{Mazumdar2019ACA} introduced a 2D detector to find important objects first, and this kind of local information can improve the performance of the overall saliency map. Recently, Chao \etal~\cite{Chao2021AMV} utilized three different FoVs in each viewport to extract rich salient features and better combined the local and global information. Furthermore, stretch weighted maps are applied in the loss function to avoid the disproportionate impact of stretching in the north and south poles of the ERP image.

(iii) ODI saliency prediction methods with multi-type inputs focus on the projection transformations of the ODIs, which has been mentioned in the Sec~\ref{sec2.1}. These methods aim to utilize the properties of different projection formats to achieve the better performance than the single ERP input~\cite{SalNet360, Djemai2020Extending2S, Chen2020SalBiNet360SP}. 
Due to the geometric distortions in the poles of ERP format, Djemai \etal~\cite{Djemai2020Extending2S} introduced a set of CP images, which are projected by five different rotational ERP images into the CNN-based approach. However, boundary-distortion and discontinuity in CP images cause the lack of global information in the extracted features. To address the problem, SalBiNet360~\cite{Chen2020SalBiNet360SP} simultaneously takes ERP and CP images as the input. It constructs a bifurcated network to predict global and local saliency maps, respectively. The final saliency output is the fusion of the global and local saliency maps. Furthermore, Zhu~\cite{Zhu2020ThePO} provided a groundbreaking multi-domain model, which decomposes the ERP image using spherical harmonics in the frequency domain and combines frequency components with multiple viewports of the ERP images in the spatial domain to extract features.

(iv) As the first to use GAN to predict the saliency maps for ODIs, SalGAN360~\cite{Chao2018Salgan360VS} provides a new generator loss, which is designed according to three evaluation metrics to fine-tune the SalGAN~\cite{Pan2017SalGANVS}. SalGAN360 constructs a different branch with the Multiple Cubic Projection (MCP) as input to simulate undistorted contents. For the attention-based learning on ODI saliency prediction, Zhu~\etal~proposed RANSP~\cite{Zhu2020RanspRA} and AAFFN~\cite{Zhu2021SaliencyPO}. Both methods contain the part-guided attention (PA) module, which is a normalized part confidence map that can highlight specific regions in the image. Moreover, an attention-aware module is introduced to refine the final saliency map. Especially, RANSP predicts the head fixations while AAFFN predicts the eye fixations.

\noindent\textbf{ODV Saliency Prediction} For the saliency prediction in ODVs, the key points are accurate saliency prediction for each frame and the temporal coherence of the viewing process. As videos with dynamic contents are widely used in real applications, deep ODV saliency prediction has received more attention in the community. Nguyen \etal~\cite{Nguyen2018YourAI} proposed a representative transfer learning framework that shifted a traditional saliency model to a novel saliency model, PanoSalNet, which is similar to~\cite{Assens2017SaltiNetSP} and~\cite{SalNet360}. By contrast, Cheng \etal~\cite{cheng2018cube} proposed a spatial-temporal network consisting of a static model and a ConvLSTM module. The static model is inspired by~\cite{Zhou2016LearningDF} and ConvLSTM~\cite{shi2015convolutional} is used to aggregate temporal information. They also implemented the Cube Padding technique to connect the cube faces by propagating the shared information across the views. Similar to~\cite{Mazumdar2019ACA}, a viewport saliency prediction model is proposed in~\cite{Qiao2021ViewportDependentSP} which first studies human attention to detect the desired viewports of the ODV and then predict the fixations based on the viewport content. Especially, the proposed Multi-Task Deep Neural Network (MT-DNN) model takes both the viewport content and location of the viewport as the input and its structure follows~\cite{cheng2018cube} which employs a CNN and a ConvLSTM to explore both spatial and temporal features. One more representative is proposed by~\cite{Zhang2018SaliencyDI}, in which the convolution kernel is defined on a spherical crown and the convolution operation corresponds to the rotation of kernel on the sphere. Considering the common planar ERP format, Zhang \etal~\cite{Zhang2018SaliencyDI} re-sampled the kernel based on the position of the sampled patches on ERP. There also exist some works based on novel learning strategies. Xu \etal~\cite{Xu2019PredictingHM} developed the saliency prediction network of head movement (HM) based on deep reinforcement learning (DRL). The proposed DRL-based head movement prediction approach owns offline and online versions. In offline version, multiple DRL workflows determines potential HM positions at each panoramic frame and generate a heat map of the potential HM positions. In online version, the DRL model will estimate the next HM position of one subject according to the currently observed HM position. Zhu \etal~\cite{Zhu2021ViewingBS} proposed a graph-based CNN model to estimate the fraction of the visual saliency via Markov Chains. The edge weights of the chains represent the characteristics of viewing behaviors, and the nodes are feature vectors from the spatial-temporal units.
\vspace{-2pt}
\subsubsection{Gaze Behavior}
\label{sec3.4.2}

Gaze following, also called gaze estimation, is related to detecting what people in the scene look at and are absorbed in. As normal perspective images are NFoV captured, gaze targets are always out of the scene. ODI gaze following is proposed to solve this problem because ODIs have a great ability to capture the entire viewing surroundings. 
Previous 3D gaze following methods can directly detect the gaze target of a human subject in the sphere space but ignore scene information of ODIs, which performs gaze following not well. Gaze360~\cite{kellnhofer2019gaze360} collects a large-scale gaze dataset using fish-eye lens rectification to pre-process the images. However, due to the distortion caused by the sphere-to-plane projection, the gaze target maybe not be in the 2D sightline of the human subject in long-distance gaze, which is no longer the same in 2D images. Li \etal~\cite{LiSGZZG21} proposed the first framework for ODI gaze following and also collected the first ODI gaze following dataset, called GazeFollow360. They detected the gaze target within a local region and a distant region. For ODI gaze prediction, Xu \etal~\cite{XuDWSSYG18} built a large-scale eye-tracking dataset for dynamic 360$^\circ$ immersive videos and gave a detailed analysis of gaze prediction. They utilized the temporal saliency, spatial saliency and history gaze path for gaze prediction with a combination of CNN and LSTM, which is similar to the architecture proposed by~\cite{wu2020spherical}.

\noindent \textbf{Challenges and potential:} ODI contains richer context information that can boost gaze behaviour understanding. However, some challenges remain. First, there are few specific gaze following and gaze prediction datasets specific for ODI. Data is the "engine" of DL-based methods, so collecting the quantitative and qualitative datasets is necessary. Second, due to the distortion problem in sphere-to-plane projection types, future research should consider how to correct this distortion via geometric transformation. Finally, both gaze following and gaze prediction in ODI need to understand wider scene information compared with normal 2D images. The spatial context relation should be further explored.

\vspace{-5pt}
\subsubsection{Audio-Visual Scene Understanding}
\label{sec3.4.3}

Because ODVs can provide the observers with an immersive understanding of the entire surrounding environments, recent research focuses on audio-visual scene understanding on ODVs. Due to its enabling viewers to experience sound in all directions, the spatial radio of ODV is an essential cue for full scene awareness. As the first work on the omnidirectional spatialization problem, Morgado \etal~\cite{Morgado2018SelfSupervisedGO} designed a four-block architecture applying self-supervised learning to generate the spatial radio, given the mono audio and ODV as the joint inputs. They also proposed a representative self-supervised framework~\cite{morgado2020learning} for learning representations from the audio-visual spatial content of ODVs. 
In~\cite{Masuyama2020SelfsupervisedNA}, ODIs combined with the multichannel audio signals are applied to localize sound source object within the visual observation. The self-supervised training method includes two DNN models: one for visual object detection and another for sound source estimation. Both DNN models are trained based on variational inference. Vasudevan \etal~\cite{vasudevan2020semantic} simultaneously achieved an audio task, spatial sound super-resolution, and two visual tasks, dense depth prediction, and semantic labeling of the scene. They proposed a cross-modal distillation framework, including a shared encoder and three task-specific decoders, to transfer knowledge from vision to audio. For the audio-visual saliency prediction on ODVs, AVS360~\cite{Chao2020TowardsAS} is the first end-to-end framework with two branches to understand audio and visual cues. Especially, AVS360 considers geometric distortion in ODV and extracts the spherical representation from the cube map images. Furthermore, as the first user behavior analysis for audio-visual content in ODV, Chao \etal~\cite{Chao2020AudioVisualPO} designed the comparative studies using ODVs with three different audio modalities and demonstrated that audio cues can improve the audio-visual attention in ODV. 

\noindent\textbf{Discussion:} Based on the above analysis, most works in this research domain process ERP images as normal 2D images and ignore the inherent distortions. Future research may explore how better combine spherical imaging characteristics and geometrical information of ODI with the spatial audio cues to provide a more realistic audio-visual experience. 

\vspace{-3pt}
\subsubsection{Visual Question Answering }
\label{sec3.4.4}


Visual question answering (VQA) is a comprehensive and interesting task that combines computer vision (CV), natural language processing (NLP), and knowledge representation $\&$ reasoning (KR). Wider FoV ODIs and ODVs are more valuable and challenging for the VQA research because they can provide stereoscopic spatial information similar to the human visual system. VQA 360$^\circ$, proposed in~\cite{chou2020visual}, is the first VQA framework on ODI. It introduces a CP-based model with multi-level fusion and attention diffusion to reduce spatial distortion. Meanwhile, the collected VQA 360$^\circ$ dataset provides a benchmark for future developments. Furthermore, Yun \etal~\cite{Yun2021PanoAVQAGA} proposed the first ODV-based VQA work, Pano-AVQA, which combines information from three modalities: language, audio, and ODV frames. The fused multi-modal representations extracted by a transformer network provide a holistic semantic understanding of omnidirectional surroundings. 
They also provided the first spatial and audio-VQA dataset on ODVs.


\noindent\textbf{Discussion and Challenges:} Based on the above analysis, there exist few works for the ODI$/$ODV-based VQA. Compared with the methods in 2D domain, the most considerable difficulty is how to leverage the spherical projection types, \eg, icosahedron and tangent images. As more than two dozen datasets and numerous effective networks~\cite{Liu2020InverseVQ} in the 2D domain have been published, future research may consider how to effectively transfer knowledge to learn more robust DNN models for omnidirectional vision.


\vspace{-4pt}
\section{Novel Learning Strategies}
\label{sec4}
\noindent\textbf{Unsupervised/Semi-supervised Learning. }
\label{sec4.1}
ODI data scarcity problem occurs due to the insufficient yet costly panorama annotations. This problem is commonly addressed by semi-supervised learning or unsupervised learning that can take advantage of abundant unlabeled data to enhance the generalization capacity. For semi-supervised learning, Tran \etal\cite{Tran2021SSLayout360SI} exploited the `Mean-Teacher' model~\cite{AnttiTarvainen2017MeanTA} for 3D room layout reconstruction by learning from the labeled and unlabeled data in the same scenario. For unsupervised learning, Djilali \etal~\cite{Abdelaziz2021Rethinking3I} proposed the first framework for ODI saliency prediction. It calculates the mutual information between different views from multiple scenes and combines contrastive learning with unsupervised learning to learn latent representations. Furthermore, unsupervised learning can be combined with supervised learning to enhance the generalization capacity. Yun \etal~\cite{yun2021improving} proposed to combine self-supervised learning with supervised learning for depth estimation, alleviating data scarcity and enhancing stability.




\noindent\textbf{GAN.}
\label{sec4.2}
To decrease the domain divergence between perspective images and ODIs, P2PDA~\cite{zhang2021transfer} and DENSEPASS~\cite{MaZYRS21} exploit the GAN frameworks and design an adversarial loss to facilitate semantic segmentation. In image generation, BIPS~\cite{oh2021bips} proposes a GAN framework to synthesize RGB-D indoor panoramas based on the arbitrary configurations of cameras and depth sensors.



\noindent\textbf{Attention Mechanism.}
\label{sec4.3}
For cross-view geo-localization, in~\cite{zhu2022transgeo}, ViT~\cite{dosovitskiy2020image} is utilized to remove uninformative image patches and enhance the informative image patches to higher resolution. This attention-guided non-uniform cropping strategy can save the computational cost, which is reallocated to informative patches to improve the performance. The similar strategy is adopted in the unsupervised saliency prediction~\cite{Abdelaziz2021Rethinking3I}. In~\cite{Abdelaziz2021Rethinking3I}, a self-attention model is employed to build spatial relationship between the two input and select the sufficiently invariant features.




\noindent\textbf{Transfer Learning.}
\label{sec4.4}
There exist a lot of works to transfer the knowledge learned from the source 2D domain to facilitate learning in the ODI domain for numerous vision tasks, \eg, semantic segmentation~\cite{DengYLLHW20} and depth estimation~\cite{Tateno2018DistortionAwareCF}.
Designing the deformable CNN or MLP on the pre-trained models from perspective images can enhance the model capability for ODIs in numerous tasks, \eg, semantic segmentation~\cite{DengYLLHW20,Tateno2018DistortionAwareCF, zhang2019orientation,Lee2019SpherePHDAC,Guerrero-ViuFDG20,Zhang2022BendingRD}, video super-resolution~\cite{liu2020single}, depth estimation~\cite{Tateno2018DistortionAwareCF}, and optical flow estimation~\cite{Xie2019EffectiveCN}. However, these methods heavily rely on the handcrafted modules, which lack the generalization capability for different scenarios. Unsupervised domain adaptation aims to transfer knowledge from the perspective domain to ODI domain by decreasing the domain gaps between the perspective images and ODIs. P2PDA~\cite{zhang2021transfer} and BendingRD~\cite{Zhang2022BendingRD} decrease domain gaps between perspective images and ODIs to effectively obtain pseudo dense labels for the ODIs.
Knowledge distillation (KD) is another effective technique that transfers knowledge from a cumbersome teacher model to learn a compact student model, while maintaining the student's performance.
However, we find that few works have applied KD for omnidirectional vision tasks. In semantic segmentation, ECANets~\cite{Yang2021CapturingOC} performs data distillation via diverse panoramas from all around the globe.

\label{sec4.5}
\noindent\textbf{Deep Reinforcement Learning (DRL).}
In saliency prediction,~\cite{MaiXu2021SaliencyPO} predicted the head fixation through DRL by interpreting the trajectories of head movements as discrete actions, which are rewarded by correct policies. Besides, in object detection, Pais \etal~\cite{Pais2019OmniDRLRP} provided the pedestrians' positions in the real world by considering the 3D bounding boxes and their corresponding distortion projections into the image. Another application for DRL is to select up-scaling factors adaptively based on the pixel density~\cite{Deng2021LAUNetLA}, which addresses the unevenly distributed pixel density in the ERP.



\label{sec4.6}
\noindent\textbf{Multi-task Learning.}
Sharing representations between the related tasks can increase the generalization capacity of the models and improve the performance on all involved tasks. MT-DNN~\cite{Qiao2021ViewportDependentSP} combines the saliency detection task with the viewport detection task to predict the viewport saliency map of each frame and improves the saliency prediction performance in the ODVs. DeepPanoContext~\cite{Zhang2021DeepPanoContextP3} empowers panoramic scene understanding by jointly predicting object shapes, 3D poses, semantic categories, and room layout. Similarly, HoHoNet~\cite{Sun2021HoHoNet3I} proposes a Latent Horizontal Feature (LHFeat) and a novel horizon-to-dense module to accomplish various tasks, including room layout reconstruction and per-pixel dense prediction tasks, \eg, depth estimation, semantic segmentation.



\vspace{-10pt}
\section{Applications}
\label{sec5}

\noindent\textbf{AR and VR.}
\label{sec5.1}
With the advancement of techniques and the growing demand of interactive scenarios, AR and VR have seen rapid development in recent years. VR aims to simulate real or imaginary environments, where a participant can obtain immersive experiences and personalized content by perceiving and interacting with the environment. With the advantage of capturing the entire surrounding environment with $360^\circ\times180^\circ$ FoV in ODIs, 360 VR/AR facilitates the development of immersive experiences.

\cite{kittel2020360} gives a detailed SWOT (namely strengths, weaknesses, opportunities, and threats) analysis of 360 VR to make sure that it is suitable to leverage the 360 VR to develop athletes' decision-making skills. Understanding human behaviors is crucial for the application of 360 VR. \cite{wu2020spherical} proposed a preference-aware framework for viewport prediction, and \cite{XuDWSSYG18} combined the history scan path with image contents for gaze prediction. In addition, to enhance the immersive experience, Kim \etal~\cite{KimHJH19} proposed a novel pipeline to estimate room acoustic for plausible reproduction of spatial audio with $360^{\circ}$ cameras. Importantly, acquiring 3D data is strongly desired in VR/AR to provide the sense of 3D. However, consumer-level depth sensors can only capture perspective depth maps, and panoramic depths need time-consuming stitching technologies. Therefore, monocular depth estimation techniques, \eg, OmniDepth~\cite{zioulis2018OmniDepthDD} and UniFuse~\cite{jiang2021unifuse}, are promising for VR/AR.

\noindent\textbf{Robot Navigation.}
\label{sec5.2}
In addition to SLAM mentioned in Sec.~\ref{sec3.3.3}, we further discuss the related applications of ODI/ODV in the field of robot navigation, including the telepresence system, surveillance, and DL-based optimization methods. 

The telepresence system aims to overcome the space constraints to enable people to remotely visit and interact with each other. ODI/ODV is gaining popularity by providing a more realistic and natural scene, especially in outdoor activities with open environments~\cite{Heshmat2018GeocachingWA}.~\cite{Zhang2018A3} proposed a prototype of an ODV-based telepresence system to support more natural interactions and the remote environment exploration, where real walking in the remote environment can simultaneously control the relevant movement of the robot platform. Surveillance aims to replace humans for security purposes, in which the calibration is vital for sensitive data. Accordingly, Pudics \etal~\cite{Pudics2015SafeRN} proposed a safe navigation system tailored for obstacle detection and avoidance with a calibration design to obtain the proper distance and direction. Compared with NFoV images, panoramic images can reduce the computational cost significantly by providing complete FoV in a single shot. Moreover, Ran \etal~\cite{Ran2017ConvolutionalNN} proposed a lightweight framework based on the uncalibrated $360^{\circ}$ cameras. The framework can accurately estimate the heading direction by formulating it into a series of classification tasks and avoid redundant computation by saving the calibration and correction processes. To address dark environments, \eg, underground mine, Mansouri \etal~\cite{Mansouri2019VisionbasedMN} presented another DNN model by utilizing online heading rate commands to avoid the collision in the tunnels and calculating depth information online within the scene.

\noindent\textbf{Autonomous Driving.}
It requires a full understanding of the surrounding environment, which omnidirectional vision excels at. Some works focus on setting up $360^{\circ}$ platform for autonomous driving~\cite{sun2019multimodal,beltran2020towards}. Specifically, \cite{sun2019multimodal} utilized a stereo camera, a polarization camera and a panoramic camera to form a multi-modal visual system to capture omnidirectional landscape.~\cite{beltran2020towards} introduced a multi-modal 360$^\circ$ perception proposal based on visual and LiDAR scanners for 3D object detection and tracking. In addition to the platform, the emergence of public omnidirectional datasets for autonomous driving are crucial for the application of DL methods. Caeser \etal~\cite{caesar2020nuscenes} were the first to introduce the relevant dataset which carries six cameras, five radars and one LiDAR. All devices are with $360^{\circ}$ FoV. Recently, OpenMP dataset~\cite{zhang2022openmpd} is captured by six cameras and four LiDARs, which contains scenes in the complex environment, \eg, urban areas with overexposure or darkness. 
Kumar \etal \cite{kumar2021omnidet} presented a multi-task visual perception network, which consists of six vital tasks in autonomous driving: depth estimation, visual odometry, senmantic segmentation, motion segmentation, object detection and lens soiling detection. Importantly, as real-time performance is crucial for autonomous driving and embedding systems in vehicles often have limited memory and computational resources, lightweight DNN models are more favored in practice.

\vspace{-8pt}
\section{Discussion and New Perspectives}
\label{sec6}
\noindent\textbf{Cons of Projection Formats.} 
ERP is the most prevalent projection format due to its wide FoV in a planar format. The main challenge for ERP is the increasing stretching distortion towards poles. Therefore, many works were proposed to design specific convolution filters against the distortion~\cite{coors2018spherenet,Su2017LearningSC}. By contrast, CP and tangent images are distortion-less projection formats by projecting a spherical surface into multiple planes. 
They are similar to the perspective images, and therefore can make full use of many pre-trained models and datasets in the planar domain~\cite{li2022omnifusion}. However, CP and tangent images suffer from the challenges of higher computational cost, discrepancy and discontinuity. 

We summarize two potential directions for utilizing CP and tangent images: (i) Redundant computational cost are resulted from large overlapping regions between projection planes. However, the pixel density varies among different sampling positions. The computation can be more efficient through allocating more resources for dense regions (\eg, equator) and less resources for sparse regions (\eg, poles) with reinforcement learning~\cite{Deng2021LAUNetLA}. (ii) Currently, different projection planes are often processed in parallel, which lacks the global consistency. To overcome the discrepancy among different local planes, it is effective to explore an additional branch with ERP as the input~\cite{wang2020bifuse} or attention-based transformers to construct non-local dependencies~\cite{li2022omnifusion}. However, these constraints are mainly added to the feature maps, instead of the predictions. Moreover, the discrepancy can be also solved from the distribution consistency of predictions, \eg, the consistent depth range among different planes and the consistent uncertainty scores for the same edges and large gradient regions.

\noindent\textbf{Data-efficient Learning.} A challenge for DL methods is the need for large-scale datasets with high-quality annotations. However, for omnidirectional vision, constructing large-scale datasets is expensive and tedious. Therefore, it is necessary to explore more data-efficient methods. One promising direction is to transfer the knowledge learned from models trained on the labeled 2D dataset to models to be trained on the unlabeled panoramic dataset. Specifically, domain adaptation approaches can be applied to narrow the gap between perspective images and ODIs~\cite{MaZYRS21}. KD is also an effective solution by transferring learned feature information from a cumbersome perspective DNN model to a compact DNN model learning ODI data~\cite{Yang2021CapturingOC}. Finally, recent self-supervised methods, \eg,~\cite{yan2022multi}, demonstrate the effectiveness of pre-training without the need of additional training annotations.


\noindent \textbf{Physical Constraint.}  Existing methods for the perspective images are limited in inferring the lighting of the global scene and unseen regions. Owing to the wide FoV of ODIs, complete surrounding environment scenes can be captured. Furthermore, the reflectance can be revealed according to the physical constraints between the lighting and scene structure based on~\cite{li2021lighting}. Therefore, a future direction can be jointly leveraging computer graphics, like ray tracing, and rendering models to help calculate reflectance, which, in turn, contributes to higher-precision global lighting estimation. Additionally, it is promising to process and render ODIs based on the lighting transportation theory.

\noindent\textbf{Multi-modal Omnidirectional Vision.} It refers to the process of learning representations from different types of modalities (\eg, text-image for visual question answering, audio-visual scene understanding) using the same DNN model. This is a promising yet practical direction for ominidirectional vision.
For instance, \cite{beltran2020towards} introduces a multi-modal perception framework based on the visual and LiDAR information for 3D object detection and tracking. However, existing works in this direction treat ODIs as the perspective images and ignore the inherent distortion in the ODIs. Future works may explore how to utilize the advantage of ODIs, \eg, complete FoV, to assist the representation of other modalities. Importantly, the acquisition of different modalities has obvious discrepancies. For example, capturing RGB images is much easier than that of depth maps. Therefore, a promising direction is to extract available information from one modality and then transfer to another modality via multi-task learning, KD, etc. However, the discrepancy among different modalities should be considered to ensure multi-modal consistency.

\noindent \textbf{Potential for Adversarial Attacks.} There exist few studies focusing on adversarial attacks towards omnidirectional vision models. Zhang \etal~\cite{zhang2022sp} proposed the first and representative attack approach to fool DNN models by
perturbing only one tangent image rendered from
the ODI. The proposed attack is sparse as it disturbs only a small part of the input ODI. Therefore, they further proposed a position searching method to search for the tangent point on the spherical surface. There are numerous promising yet challenging research problems in this direction, \eg, analyzing the generalization capacity of attacks among different DNN models for ODIs, white-box attacks for network architectures and training methods, and defenses against attacks. 

\noindent \textbf{Potential for Metaverse.} Metaverse aims to create a virtual world containing large-scale high-fidelity digital models, where users can freely create contents and obtain immersive interactive experience. Metaverse is facilitated by the AR and VR headsets, in which ODIs are favored due to the complete FoV. Therefore, a potential direction is to generate high-fidelity 2D/3D models from ODIs and simulate the real-world objects and scenes in great details. In addition, to help users obtain immersive experience, techniques that analyze and understand human behavior (\eg, gaze following, saliency prediction) can be further explored and integrated in the future.

\noindent \textbf{Potential for Smart City.} Smart city focuses on collecting data from the city with various devices and utilizing information from the data to improve efficiency, security and convenience, etc. Taking advantage of the characteristics of ODI in street-view images can facilitate the development of urban forms comparison. As mentioned in Sec. \ref{sec3.1.2}, a promising direction is to convert street-view images into satellite-view images for urban planning. Except for room layout discussed in Sec. \ref{sec3.3.1}, ODIs can also be applied in more interior designs. To achieve floorplan design, Wang \etal~\cite{wang2021actfloor} leveraged human-activity maps and editable furniture placements to improve the interaction with users. However, the input of~\cite{wang2021actfloor} is the boundary of the exterior wall, resulting in limitation of the visualization and manipulation. Future works might consider operating directly on the ODIs to make the interior design observable in all directions, boosting the development of interaction and making professional service accessible.

\vspace{-0.25cm}
\section{Conclusion}
\label{sec7}


In this survey, we comprehensively reviewed and analyzed the recent progress of DL methods for omnidirectional vision. We first introduced the principle of omnidirectional imaging, convolution methods and datasets. We then provided a hierarchical and structural taxonomy of the DL methods. For each task in the taxonomy, we summarized the current research status and pointed out the opportunities and challenges. We further provided a review of the novel learning strategies and applications. After constructing connections among current approaches, we discussed the pivotal problems to be solved and indicated promising future research directions. We hope this work can provide some insights for researchers and promote progress in the community.




\ifCLASSOPTIONcaptionsoff
  \newpage
\fi

\bibliographystyle{IEEEtran}
\bibliography{reference}

\begin{IEEEbiography}[{\includegraphics[width=1in,height=1.2in,clip,keepaspectratio]{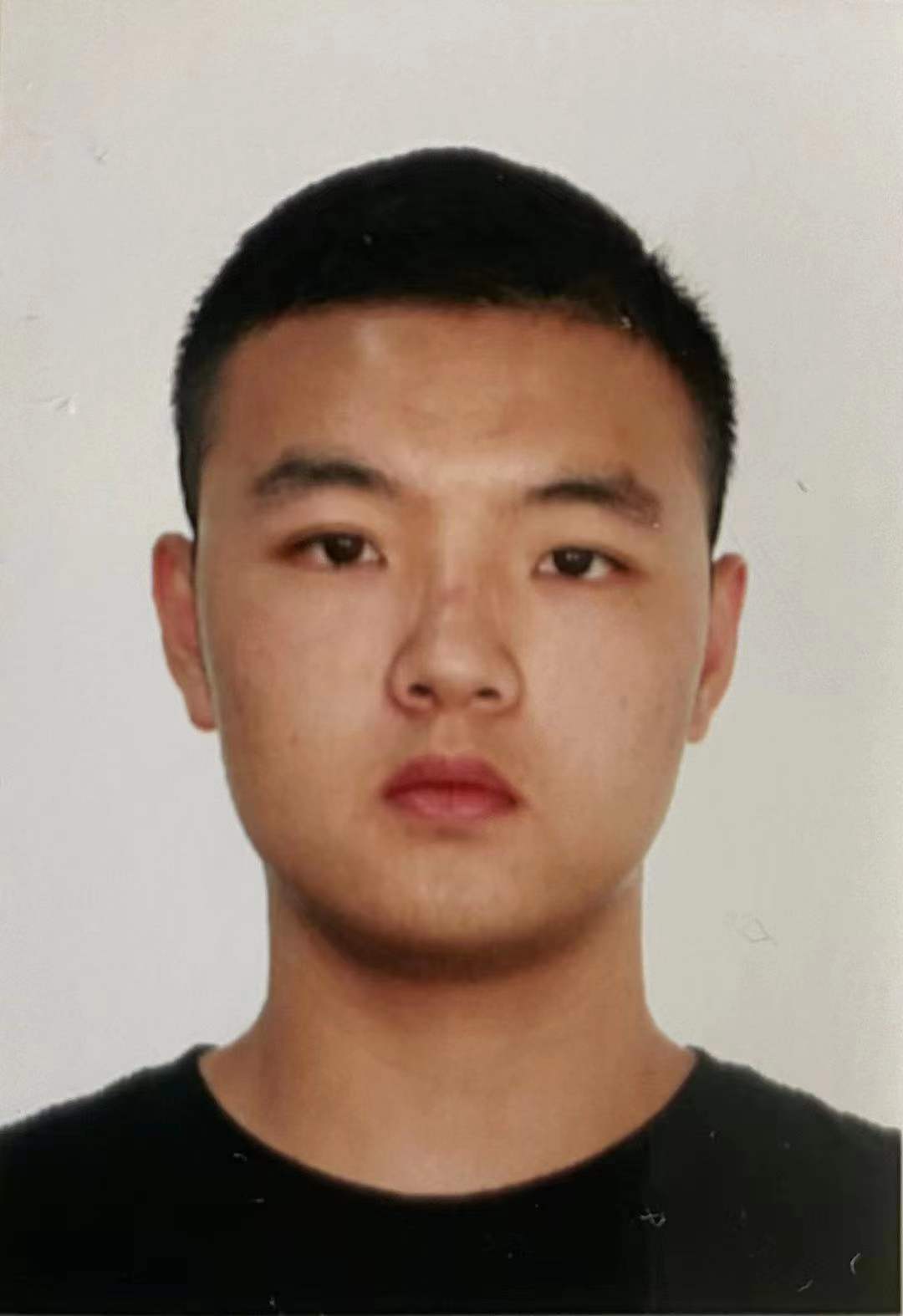}}] {Hao Ai}
is a Ph.D. student in the Visual Learning and Intelligent Systems Lab,  Artificial Intelligence
Thrust,  Guangzhou Campus, The Hong Kong University of Science and Technology (HKUST). His research interests include pattern recognition (image classification, face recognition, etc.), DL (especially uncertainty learning, attention, transfer learning, semi- /self-unsupervised learning), omnidirectional vision.
\vspace{-0.35cm}
\end{IEEEbiography}

\begin{IEEEbiography}[{\includegraphics[width=1in,height=1.2in,clip,keepaspectratio]{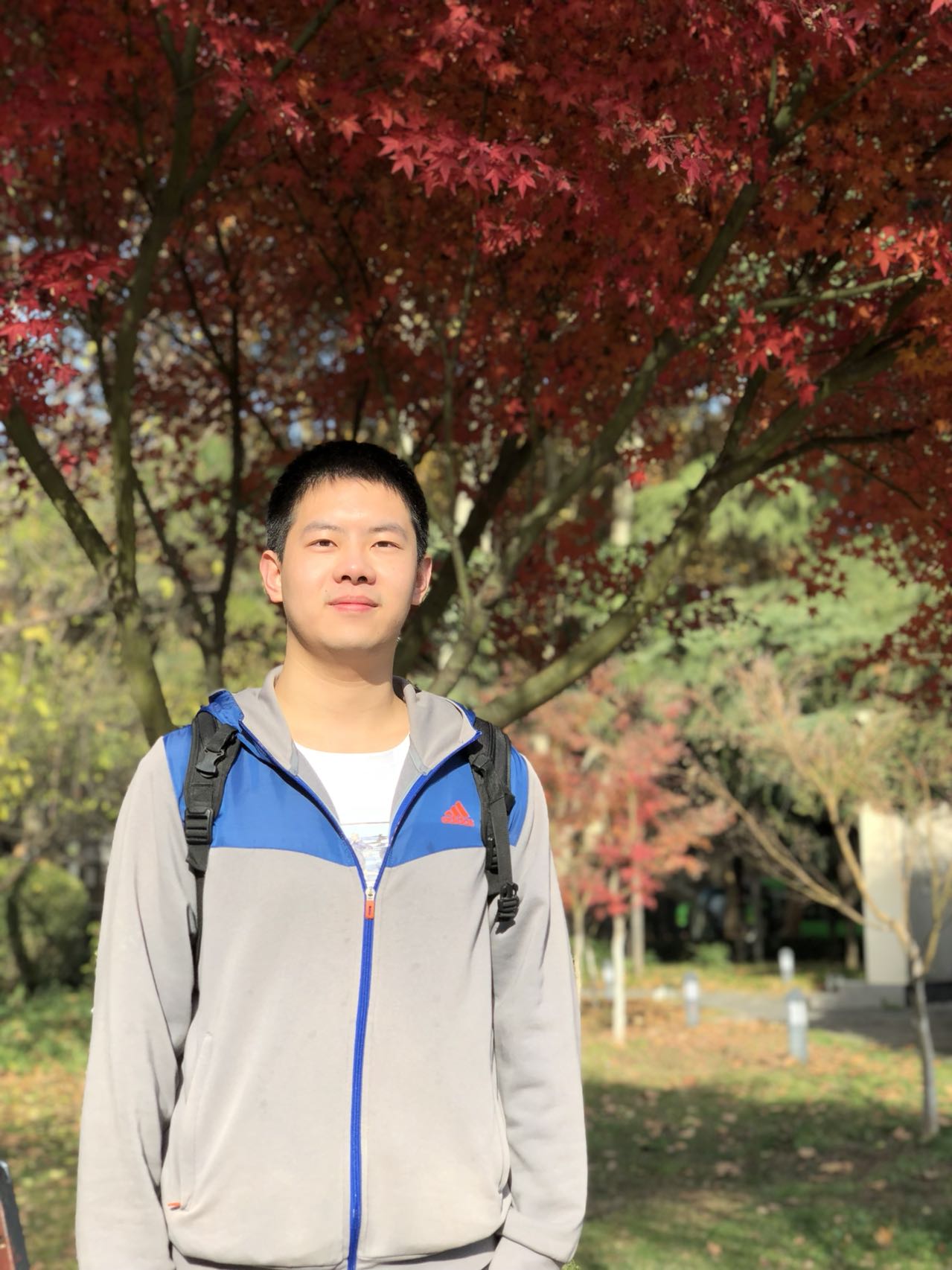}}] {Zidong Cao}
is a research assistant in the Visual Learning and Intelligent Systems Lab,  Artificial Intelligence
Thrust,  Guangzhou Campus, The Hong Kong University of Science and Technology (HKUST). His research interests include 3D vision (depth completion, depth estimation, etc.), DL (self-supervised learning, weakly-supervised learning, etc.) and omnidirectional vision.
\vspace{-0.35cm}
\end{IEEEbiography}
\begin{IEEEbiography}[{\includegraphics[width=1in,height=1.2in,clip,keepaspectratio]{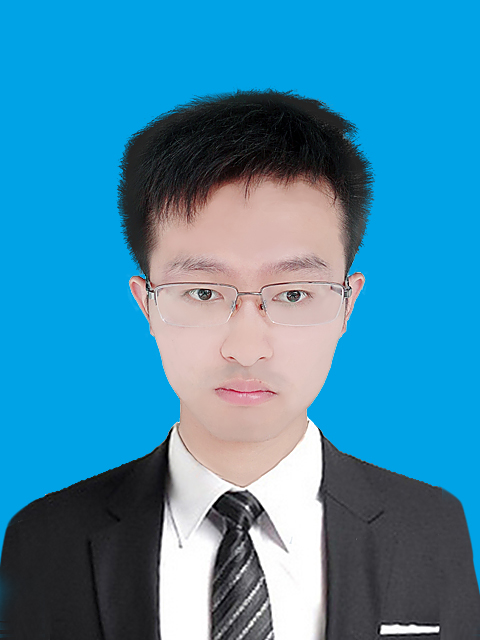}}]{JinJing Zhu}
is a Ph.D. student in the Visual Learning and Intelligent Systems Lab, Artificial Intelligence Thrust, Guangzhou Campus, The Hong Kong University of Science and Technology (HKUST). His research interests include CV (image classification, person re-identification, action recognition, etc.), DL (especially transfer learning, knowledge distillation, multi-task learning, semi-/self-unsupervised learning, etc.), omnidirectional vision, and event-based vision.
\vspace{-0.35cm}
\end{IEEEbiography}
\begin{IEEEbiography}[{\includegraphics[width=1in,height=1.2in,clip,keepaspectratio]{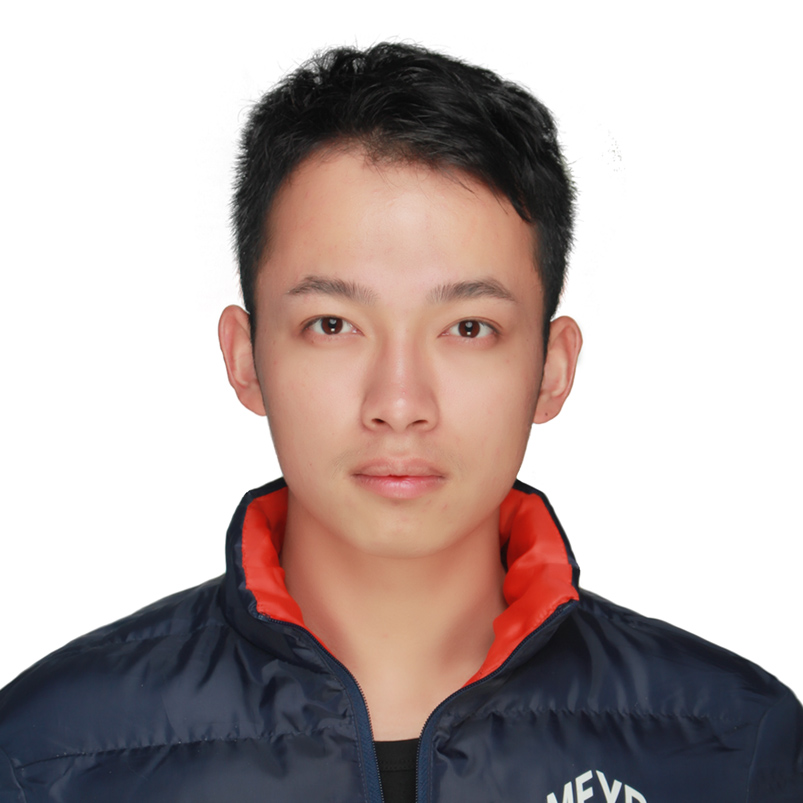}}]{Haotian Bai}
is a research assistant at Visual Learning and Intelligent Systems Lab, AI Thrust, Information Hub, Guangzhou Campus, The Hong Kong University of Science and Technology (HKUST). His research interests include 3D reconstruction (e.g., human and room), CV(especially attention, unsupervised/weakly supervised learning, and domain adaptation), and causal inference.
\vspace{-0.35cm}
\end{IEEEbiography}
\begin{IEEEbiography}[{\includegraphics[width=1in,height=1.2in,clip,keepaspectratio]{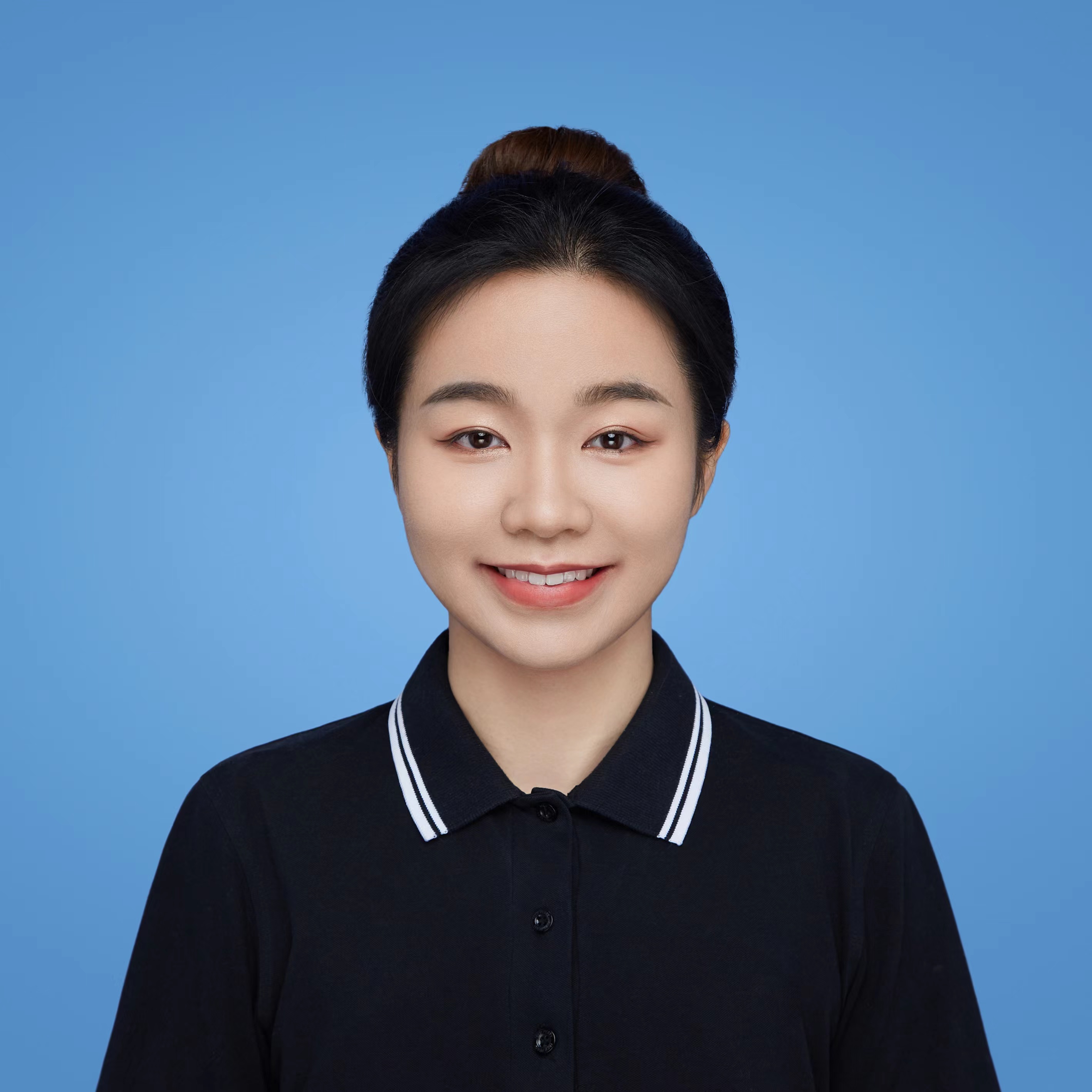}}]{Yucheng Chen}
is a Mphil student in the Visual Learning and Intelligent Systems Lab,  Artificial Intelligence Thrust,  Guangzhou Campus, The Hong Kong University of Science and Technology (HKUST). Her research interests include low-level vision (neural rendering, reflection removal, etc.), DL (especially domain adaptation, transfer learning, semi- /self-unsupervised learning), omnidirectional vision.
\vspace{-0.35cm}
\end{IEEEbiography}
\begin{IEEEbiography}[{\includegraphics[width=1in,height=1.2in,clip,keepaspectratio]{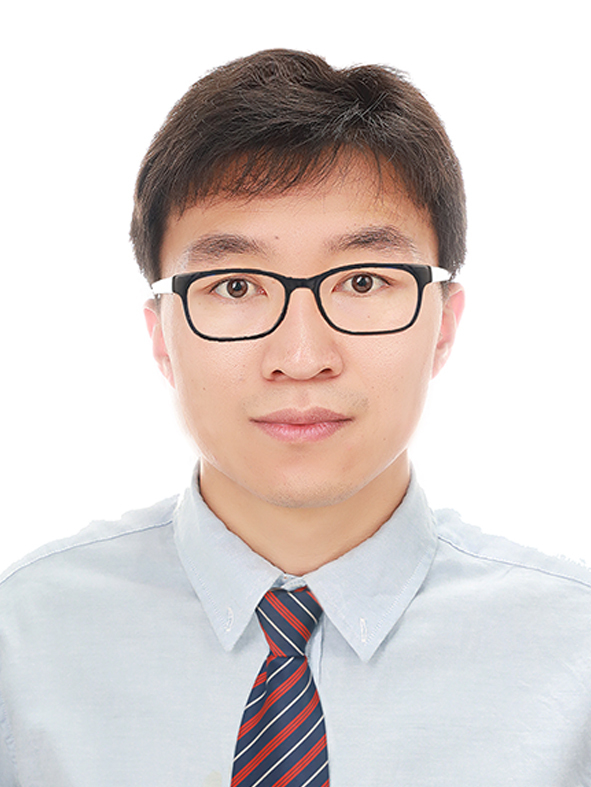}}] {Lin Wang} is an assistant professor in the Artificial Intelligence Thrust, HKUST GZ Campus, HKUST Fok Ying Tung Research Institute, and an affiliate assistant professor in the Dept. of CSE, HKUST, CWB Campus. He got his PhD degree with honor from Korea Advanced Institute of Science and Technology (KAIST).  His research interests include computer vision/graphics, machine learning and human-AI collaboration.
\end{IEEEbiography}

    
    

\ifarXiv
    

\section*{Supplementary material}

# Deep Learning for Omnidirectional Vision: A Survey and New Perspectives
# –Appendix–

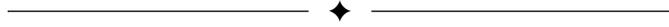

## 1 COMPLETE DATASETS OF ODI AND ODV

The dataset is important for the development of DL methods on ODI and ODV. With the emergence of the spherical imaging devices and stitching strategies, quantitative and qualitative $360°$ datasets are becoming publicly available. Some are collected from the real world, while some are synthesized by different technologies, *e.g.*, computer graphics and rendering. Table. 1 shows the commonly used datasets in various vision tasks, which are various in size, resolution, data source, etc.

In Table. 1, the ODI datasets are listed in the first 34 rows, which cover both the real-world and synthetic types. Regarding the real-world ODI datasets, most of them, *e.g.*, [1], [2], [3], [4], [5], [6], [7], are collected from indoor scenes to facilitate the tasks ranging room layout reconstruction, object detection to depth estimation. However, real-world optical flow and gaze flow datasets are comparably limited due to the difficult process of acquisition. Additionally, Most real-world datasets only provide images of 2D projection modality and are applied on some specific task. Conversely, [1] contained three modalities, including 2D, 2.5D, and 3D, where multiple modalities are beneficial for the cross-modal learning. Regarding the synthetic ODI datasets, images are high-quality with less noise, and the abundant labels are easier to obtain during the generation process. For instance, SUNCG [8] is created via the Plannar5D platform, and all the 3D scenes are composed of individually labeled 3D object meshes. Structured3D [3] and OmniFlow [9] utilize the rendering engine to generate photo-realistic images, including rich 3D structure annotations and corresponding optical flow, respectively. In addition, there also exist some datasets, *e.g.*, Omni-SYNTHIA [10], extracted from the large synthetic ones for specific tasks.

In the latter 17 rows of Table. 1, we list the ODV datasets. ODV data can be regarded as a sequence of ODI frames, and are mainly collected from public video sharing platforms like Vimeo and YouTube. Currently, most ODV datasets are downloaded or obtained in the real world, instead of synthesizing. In Table. 1, there are several datasets for the VQA task, while there are less datasets for other tasks.

## 2 COMPARISON OF THE REPRESENTATIVE METHODS

### 2.1 Comparison of Semantic Segmentation

As shown in Table. 4, we compare the representative methods of semantic segmentation tasks and directly utilize the results from their original papers for fair comparison. Moreover, we list the details of these methods, such as the input formats, the training backbones, whether they use deformable networks and their training methods. We mainly discuss the popular benchmark datasets Stanford2D3D [1], DensePASS [11], and SUN360 [12]. From these results, we have some key observations: (i) Compared with ERP [13], [14], icosahedron projection [15], [16] has resolved the spherical distortion well, and we can increase the number of subdivisions further to mitigate the spherical distortion in the future work; (ii) With the success of the Transformer, transformer-based methods [14] can achieve better performance than CNN-based methods [13]. (iii) From the comparison between Ma *et al.* [17], Zhang *et al.* [18], and Zhang *et al.* [14], we find that designing the deformable CNN or MLP on the pre-trained models from 2D images can enhance the modeling capability of ODIs with significant distortions and then improve the performance of ODIs tasks; (iv) Unsupervised domain adaptation can alleviate the burdens of high-quality annotations through transferring knowledge from the labeled source domain to the unlabeled target domain. (v) We also show the visual results on the representative ODI methods as shown in Figure. 3, 4, 5, 6, 7, 8, 9.

### 2.2 Comparison of Monocular Depth Estimation

The experimental results of representative methods for monocular depth estimation are presented in Table. 2 and Fig. 2, due to the limitation of space in the main paper.

We adopt commonly used metrics in previous methods [19], [20], including Mean Absolute Error (MAE), Absolute Relative Error (Abs Rel), Root Mean Square Error (RMSE), and accuracy within a threshold $i$, where $i \in \{1.25, 1.25^2, 1.25^3\}$. Parameters and FPS of models tested on Stanford2D3D dataset [1] are also presented in Table. 2. The commonly adopted image resolutions of Stanford2D3D [1], Matterport3D [2] and 3D60 [21] datasets are $512 \times 1024$, $512 \times 1024$, and $256 \times 512$, respectively.

From Table. 2, we can find that multiple projection formats [22], [23] and tailored network structures [19] can both improve the performance. From the overall results, OmniFusion [20] outperforms other methods in most metrics. As shown in Fig. 2(b), OmniFusion predicts clearer structures than UniFuse. We ascribe it into two main reasons: (i) The utilization of tangent images which mitigate distortions significantly. (ii) The superiority of attention-based transformers to construct non-local dependencies and capture global context to enhance the structure of the scene. Although the model of OmniFusion [20] is relatively compact, the transformation between tangent and ERP formats require

much time, resulting in low efficiency. By contrast, UniFuse [23] and HoHoNet [24] are two compact models, which can both achieve real-time speed (over 30 FPS). UniFuse [23] exploits an unidirectional fusion mechanism, which outperforms BiFuse [22] in both efficiency and accuracy. HoHoNet improves the efficiency by flattening the features along the vertical directions. As shown in Fig. 2(a), the visual results on real-world datasets also demonstrate their effectiveness. For different datasets, synthesized 3D60 [21] dataset is relatively easier than the real-world Stanford2D3D [1] and Matterport3D datasets [2].

### 2.3 Comparison of Room Layout Reconstruction

We further compare the representative studies in room layout estimation from both quantitative and qualitative perspectives. As shown in Table. 3, we list the experimental results on Matterport3D [2] dataset to give a complete comparison on both 3D, 2D IOU and FPS. From the overall results, Led2-net [25] obtains the best performance over other frameworks, which proves the effectiveness of formulating the task as depth prediction on the horizontal line of panorama. However it's worth noticing that AltantaNet [26] outperforms Led2-Net [25] on both 2D and 3D IOU in scenarios with more corners. It demonstrates that AltantaNet [26] is not only lightweight but also superior on complex situations. The advantage is presented vividly at the first two rows of Fig. 1, showing that that AtlantaNet [26] produces more details and handles more difficult situations.

Furthermore, we combine the quantitative data and collected visual results from other frameworks to display how the field is developing. For instance, earlier works like [27], [28] gives satisfactory results on the simple layouts, but are impotent to address layouts with more corners. What's more, distortion is not well handled that leaves some of results unpleasant. Later works like [25], [26], [29] provides more efficient data encoding or modeling methodology that fuels the both the speed and the accuracy on more challenging tasks.

### 2.4 Comparison of Saliency Prediction

The experimental studies of the representative methods for saliency prediction are provided in Table. 5, Table. 6 and Fig. 10 due to the limitation of space in the main paper. For ODI and ODV saliency prediction, there exist several common evaluation metrics: Kullback-Leibler divergence (KLD), linear correlation coefficient (CC), normalized scan path saliency (NSS), area under ROC curve (AUC), similarity (SIM), and the variants of AUC, *e.g.*, shuffled AUC (sAUC), AUC-Judd (AUC-J), AUC-Borji(AUC-B) (more details about these saliency metrics can be viewed in [30]).

From these results, there are several key observations. (i) For the ODI saliency prediction, there are three main evaluation benchmarks, including two benchmarks from Salient360 Grand Challenges at ICME17 and ICME18 [31] and a real-world dataset [32]. In the ICME17 challenge benchmark, the method based on CP and dilated convolutional layers [33] achieves better performance than the method with GAN based on the viewport images [34]. Although the viewport images are cropped with narrow FoV from the ERP, there still exists the distortion in the cropped images, which may degrade the performance. Moreover, we can see that the distances between the unsupervised learning method [35] and the other two supervised methods are limited. So future research may consider improving the performance of unsupervised learning methods for ODI saliency prediction due to the lack of annotated labels. (ii) From Table. 6, we can see the rich diversity in ODV saliency prediction evaluation datasets, and it is difficult to compare the performance of these methods. Constructing a large-scale benchmark could provide a straight comparison for these ODV saliency prediction methods. (iii) Last but not least, from the visual results shown in Fig. 10, we can see that there is still a large gap between the ground truths and predicted results of these methods. Especially, compared with the discrete saliency points in ground truths, the predictions are more continuous and blurring among boundaries. More geometric information of the sphere structure and attention mechanism could be considered to improve the performance of pixel-level saliency prediction.

JOURNAL OF LATEX CLASS FILES, VOL. 14, NO. 8, AUGUST 2015    3TABLE 1: Complete summary of ODI image and video datasets. N/A indicates 'not available' and GT indicates 'ground truth'.

| Dataset | Size | Data Type | Resolution | GT | Purpose |
|---|---|---|---|---|---|
| Stanford2D3D [1] | 70496 RGB+1413 ERP images | Real | 1080×1080 | ✓ | Object Detection, Scene Understanding |
| Matterport3D [2] | 194400 RGB-D images | Real | N/A | ✓ | Object Detection, Scene Understanding |
| Structured3D [3] | 196k images | Synthetic | 512×1024 | | Object Detection, Scene Understanding, Image Synthesis, 3D Modeling |
| Salient360 [32] | 60 images | Real | 5376×2688, 18332×9166 | ✓ | Saliency Prediction |
| Salient360! [36] | 60 scenes | Real | 5376×2688, 18332×9166 | ✓ | Saliency Prediction |
| Saliency in VR [37] | 22 scenes | Real | 8192×4096 | ✓ | Saliency Prediction |
| PanoContext [4] | 700 panoramas | Real | N/A | ✓ | Room Layout Estimation |
| SUN360 [12] | 67538 images | Real | 9104×4552 | ✓ | Place Categorization and Viewpoint Recognition |
| SUNCG [8] | 45622 scenes | Synthetic | N/A | | Depth Estimation |
| Gaze360 [38] | 238 subjects | Real | N/A | ✓ | Gaze Estimation |
| OmniFlow [9] | 23,653 image pairs | Synthetic | 2048×2048 | | Optical Flow |
| omni-SYNTHIA [10] | 2269 images | Synthetic | 640×1024 | | Semantic Segmentation |
| CVIQD208 [39] | 544 images | Synthetic | 4096×2048 | | IQA |
| Pano-RSOD [40] | 9402 images | Synthetic | 2048×1024 | | Object Detection |
| Refer360 [41] | 17137 sequences | Real | 9104×4552 | ✓ | Expression Recognition |
| GazeFollow360 [42] | 10058 images | Real | 4K | ✓ | Gaze Following |
| ODI-SR Database [43] | >800 images | Real | 4K ∼ 12K | ✓ | Super Resolution |
| Zillow Indoor Dataset [44] | 71474 panoramas | Real | 1024×512 | | Room Layout Estimation |
| VOC360 [45] | 21755 images | Synthetic | N/A | | Object Detection |
| COCO-Men [45] | 7000 images | Synthetic | N/A | | Object Detection |
| UrbanCity360 [46] | 7920 images | Synthetic | N/A | | View Synthesize |
| Omniscape360 [47] | 14 weather conditions | Synthetic | N/A | ✓ | Semantic Segmentation, Depth Estimation |
| DeepPanoContext Dataset [48] | 1500 panoramas | Synthetic | N/A | ✓ | Scene Understanding |
| DensePASS [11] | 19 classes | Real | 2048×400 | | Semantic Segmentation |
| 360-indoor [5] | 3335 images | Real | 960×1920 | ✓ | Object Detection |
| Replica Dataset [49] | 18 scenes | Synthetic | N/A | | Depth Estimation |
| AOI [50] | 600 ODIs | Real | N/A | | Saliency Prediction |
| Realtor360 [51] | 2573 images | Real | N/A | | Room Layout Estimation |
| 360D Dataset [21] | 11118 houses | Synthetic | 512×256 | ✓ | Depth Estimation |
| ERA [52] | 903 frames | Real | 4K | ✓ | Object Detection |
| PASS Dataset [53] | 1050 images | Real | 6000×4000, 2048×692 | | Semantic Segmentation |
| FlyingCars [54] | 6000 images | Synthetic | N/A | | Object Detection |
| Pano3DLayout [6] | 106 scenes | Real | N/A | ✓ | Room Layout Estimation |
| Shanghaitech-Kujiale Indoor 360 [55] | 3550 360° images | Synthetic | N/A | | Depth Estimation |
| FDDB-360 [56] | 17052 images | Real | N/A | ✓ | Face Detection |
| 360-Sport [57] | 342 360° videos | Real | N/A | ✓ | Visual Pilot |
| Wild-360 [58] | 85 360° videos | Real | N/A | ✓ | Video Saliency |
| 360 Head Movement [59] | 7 scenes | Real | 3840×2048 | ✓ | Head Movement Navigation |
| VQA-OV [60] | 600 sequences | Real | 3840×1920, 7680×3840 | ✓ | Head Movement Eye Movement |
| video saliency dataset [61] | 104 360° videos | Real | N/A | ✓ | Saliency Detection |
| PVS-HM Dataset [62] | 76 videos | Real | 3K ∼ 8K | ✓ | Saliency Detection, Virtual Cinematography |
| YouTube/Vimeo [63] | 115 360° videos | Real | N/A | ✓ | Highlight Detection |
| PanoSUNCG [7] | 25k 360° images | Real | N/A | ✓ | Depth Estimation |
| VQA 360 Dataset [64] | 17000 IQA triplets | Real | 1024×512 | ✓ | VQA |
| Pano-AVQA [65] | 51.7K QA Pairs | Real | N/A | ✓ | AVQA |
| Pano2Vid [66] | 86 360° videos | Real | N/A | | Automatic Cinematography |
| Narrated 360 [67] | 864 videos | Real | 720×1280 | | Visual Grounding |
| IVQAD 2017 [68] | 160 videos | Real | 4096×2048 | | VQA |
| VR-VQA [69] | 48 sequences | Real | 3K ∼ 8K | | VQA |
| Depth360 [70] | 30000 | Real | 1.03Mpx | ✓ | Depth Estimation |
| SHD360 [71] | 6268 frames | Real | 4K | ✓ | Sound-Objection Detection |
| ASOD60K [72] | 62455 frames | Real | 4K | ✓ | Sound-Objection Detection |

TABLE 2: The quantitative results for monocular depth estimation on Stanford2D3D [1], Matterport3D [2], 3D60 datasets [21].

| Datasets | Methods | MAE ↓ | Abs Rel ↓ | RMSE ↓ | $\delta_1$ ↑ | $\delta_2$ ↑ | $\delta_3$ ↑ | Parameters | FPS |
|---|---|---|---|---|---|---|---|---|---|
| Stanford2D3D [1] | RectNet [21] | 0.3743 | 0.1996 | 0.6152 | 68.77 | 88.91 | 95.78 | - | - |
| | BiFuse [22] | 0.2343 | 0.1209 | 0.4142 | 86.60 | 95.80 | 98.60 | 253.1M | 0.9 |
| | UniFuse [23] | 0.2082 | 0.1114 | 0.3691 | 87.11 | 96.64 | 98.82 | **30.3M** | **31** |
| | SliceNet [73] | **0.1757** | 0.0995 | 0.3509 | 90.29 | 96.26 | 98.44 | 75.3M | 13 |
| | HoHoNet [24] | 0.2027 | 0.1014 | 0.3834 | 90.54 | 96.93 | 98.86 | 49.5M | 52 |
| | ACDNet [19] | 0.1870 | 0.0984 | **0.3410** | 88.72 | 97.04 | 98.95 | 87.0M | 11 |
| | OmniFusion (2-iter) [20] | - | **0.0950** | 0.3474 | **89.88** | **97.69** | **99.24** | 42.3M | 4.6 |
| Matterport3D [2] | RectNet [21] | 0.4838 | 0.2901 | 0.7643 | 68.30 | 87.94 | 94.29 | - | - |
| | BiFuse [22] | 0.3470 | 0.2048 | 0.6259 | 84.52 | 93.19 | 96.32 | - | - |
| | UniFuse [23] | 0.2814 | 0.1063 | 0.4941 | 88.97 | 96.23 | 98.31 | - | - |
| | SliceNet [73] | 0.3296 | 0.1764 | 0.6133 | 87.16 | 94.83 | 97.16 | - | - |
| | HoHoNet [24] | 0.2862 | 0.1488 | 0.5138 | 87.86 | 95.19 | 97.71 | - | - |
| | ACDNet [19] | **0.2670** | 0.1010 | 0.4629 | 90.00 | 96.78 | 98.76 | - | - |
| | OmniFusion (2-iter) [20] | - | **0.0900** | **0.3474** | **91.89** | **97.97** | **99.31** | - | - |
| 3D60 [21] | RectNet [21] | - | 0.0702 | 0.2911 | 95.74 | 99.33 | 99.79 | - | - |
| | BiFuse [22] | 0.1143 | 0.0615 | 0.2440 | 96.99 | 99.27 | 99.69 | - | - |
| | UniFuse [23] | **0.0996** | 0.0466 | 0.1968 | 98.35 | 99.65 | 99.87 | - | - |
| | OmniFusion (2-iter) [20] | - | **0.0430** | **0.1808** | **98.59** | **99.69** | **99.89** | - | - |

TABLE 3: The quantitative experimental results of room Layout estimation on Mattertport3D [74].

| Method | Backbone | 2D IOU(%) | | | | | 3D IOU(%) | | | | | FPS |
|---|---|---|---|---|---|---|---|---|---|---|---|---|
| | | overall | 4 | 6 | 8 | 10+ | overall | 4 | 6 | 8 | 10+ | |
| LayoutNet [28] | ResNet-34 | 78.73 | 84.61 | 75.02 | 69.79 | 65.14 | 75.82 | 81.35 | 72.33 | 67.45 | 63.00 | 46 |
| DuLaNet [27] | ResNet-50 | 78.82 | 81.12 | 82.69 | 74.00 | 66.12 | 75.02 | 77.02 | 78.79 | 71.03 | 63.27 | 34 |
| HorizonNet [29] | ResNet-50 | 81.71 | 84.67 | 84.82 | 73.91 | 70.58 | 79.11 | 81.88 | 82.26 | 71.78 | 68.32 | 31 |
| AtlantaNet [26] | ResNet-50 | 82.09 | 84.42 | 83.85 | 76.97 | **73.19** | 80.02 | 82.09 | 82.08 | 75.19 | **71.62** | 5 |
| HoHoNet [75] | ResNet-34 | 82.32 | 85.26 | 84.81 | 75.59 | 70.98 | 79.88 | 82.64 | 82.16 | 73.65 | 69.26 | **110** |
| Led2-net [25] | ResNet-50 | **83.91** | **86.91** | **85.53** | **78.72** | 71.79 | **81.52** | **84.22** | **83.22** | **76.89** | 70.09 | - |

TABLE 4: Qualitative comparison of semantic segmentation. The accuracy is computed as Intersect over Union (%). "S": supervised, "U": Unsupervised, "D": domain adaptation.

| Dataset | Method | backbone | Input | Deformable | Supervision | Accuracy |
|---|---|---|---|---|---|---|
| Stanford | Tateno [13] | CNNs | ERP | ✓ | U | 34.6 |
| | Zhang [14] | Transformer | ERP | ✓ | U | 51.2 |
| | Lee [15] | CNNs | Tangent | ✓ | S | 51.4 |
| | Zhang [16] | U-Net | Tangent | ✓ | S | **58.6** |
| DensePASS | Ma [17] | ResNet | ERP | ✗ | U | 39.8 |
| | Zhang [18] | ResNet | ERP | ✗ | U | 42.0 |
| | Zhang [14] | Transformer | ERP | ✓ | U | **56.4** |
| SUN360 | Viu [76] | ResNet | ERP | ✓ | S | 54.4 |

TABLE 5: Qualitative comparison of ODI saliency prediction. ∗ means that the results are not from the complete testing data.

| Dataset | Method | KLD ↓ | CC ↑ | NSS ↑ | AUC ↑ | AUC_J ↑ | SIM ↑ |
|---|---|---|---|---|---|---|---|
| ICME17 [31] | Dai [33] | **0.317** | **0.686** | **0.983** | 0.748 | - | - |
| | Chao [34] | 0.363 | 0.662 | 0.978 | - | 0.747 | - |
| ICME18 [31] | Dai [33] | 0.745 | **0.656** | **1.685** | - | 0.807 | 0.642 |
| | Chao [34] | **0.726** | 0.653 | 1.646 | - | **0.829** | **0.644** |
| | Abdelaziz [35] | 0.849 | 0.584 | 1.601 | - | 0.769 | 0.591 |
| Salient360 [32] | Lv∗ [77] | 0.428 | 0.589 | 0.945 | 0.736 | - | 0.686 |
| | Xu∗ [78] | 0.366 | 0.757 | 0.893 | 0.708 | - | - |

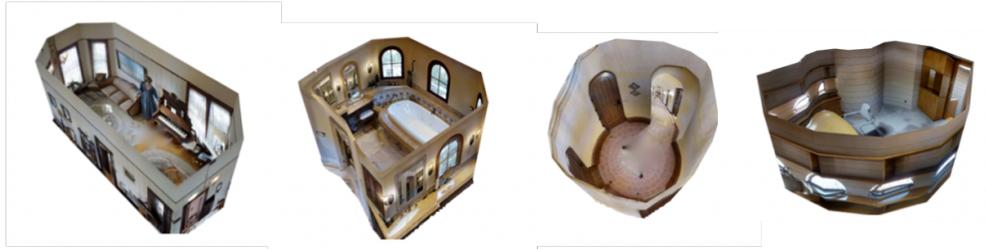

(a) AtlantaNet [26]

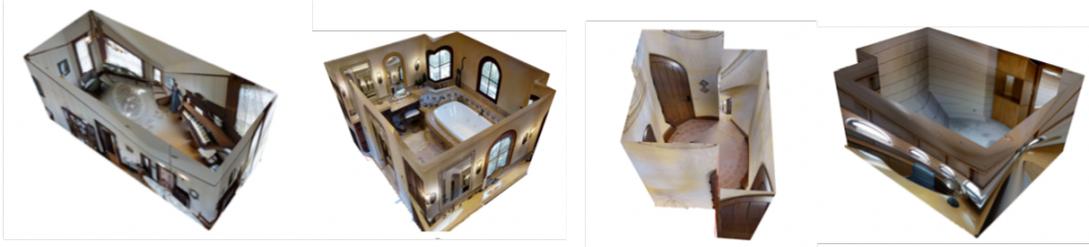

(b) HorizonNet [29]

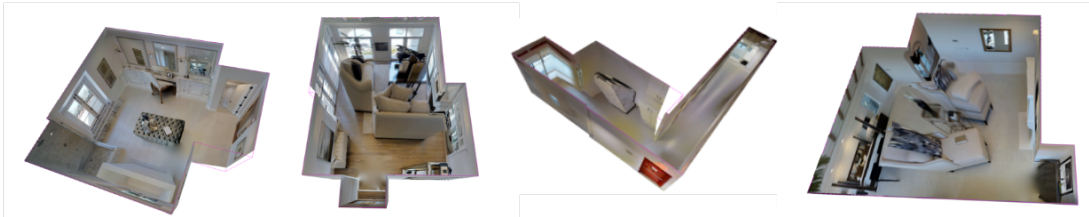

(c) SSLayout360 [84]

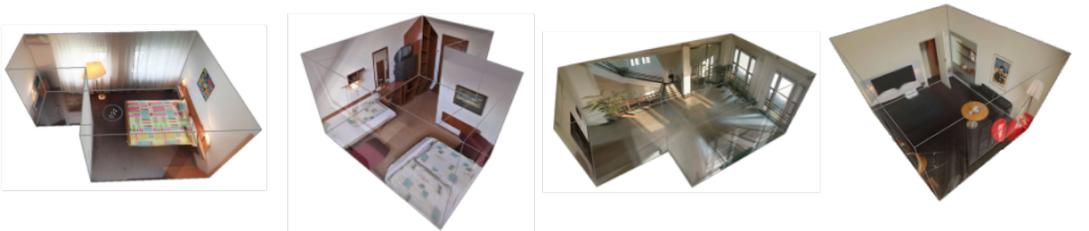

(d) LayoutNet [28]

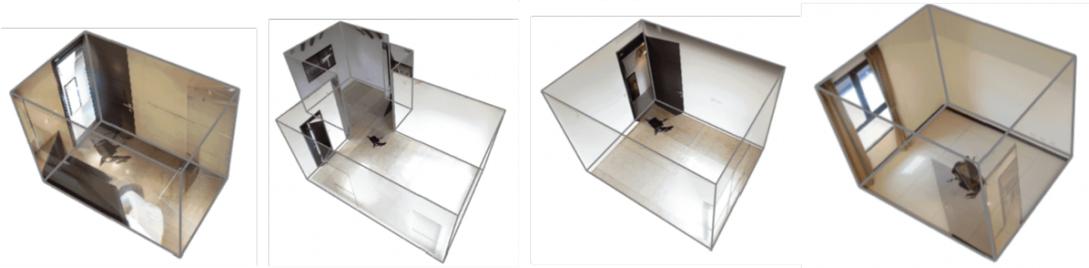

(e) DulaNet [27]

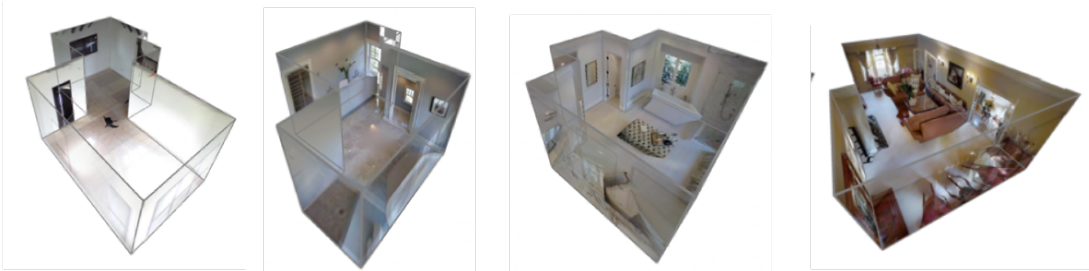

(f) Led2-net [27]

Fig. 1: Visual result on representative studies of room layout estimation.

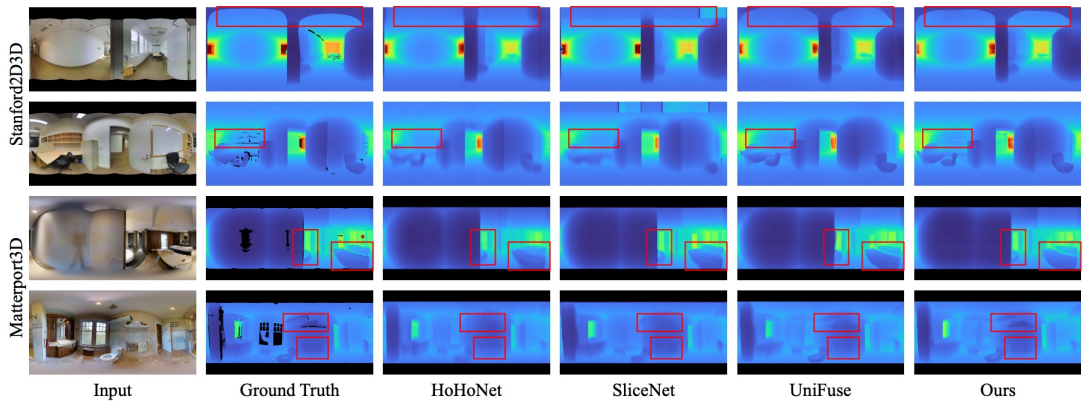

(a) Results on Stanford2D3D [1] and Matterport3D [2] datasets. Image is originally shown from [19].

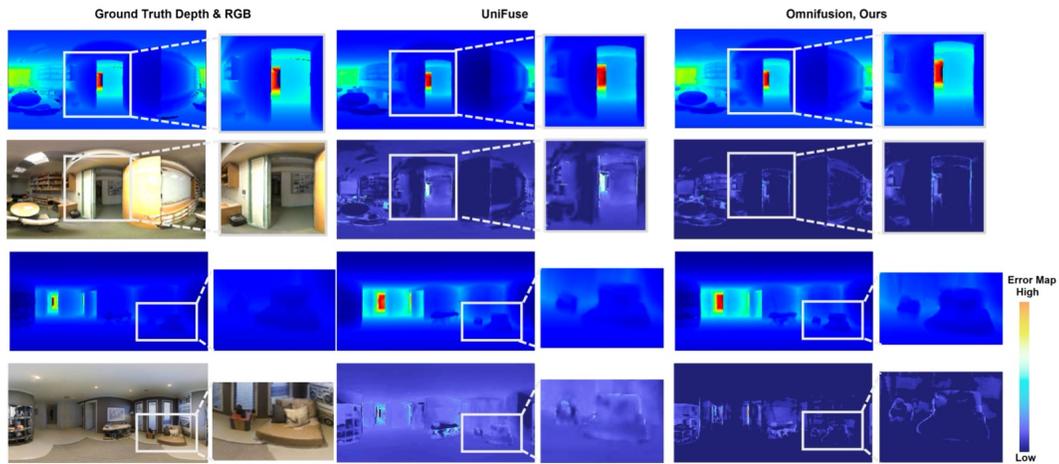

(b) Results on 3D60 [21] dataset. Image is originally shown from [20].

Fig. 2: Visual results of representative methods on monocular depth estimation.

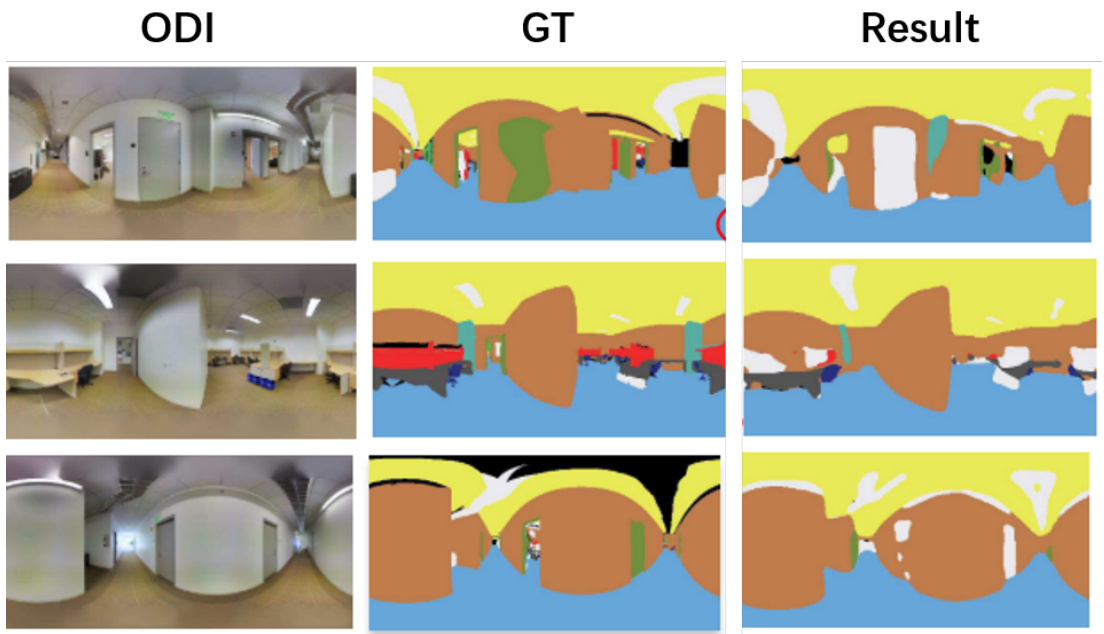

Fig. 3: Tateno *et al*. [13]

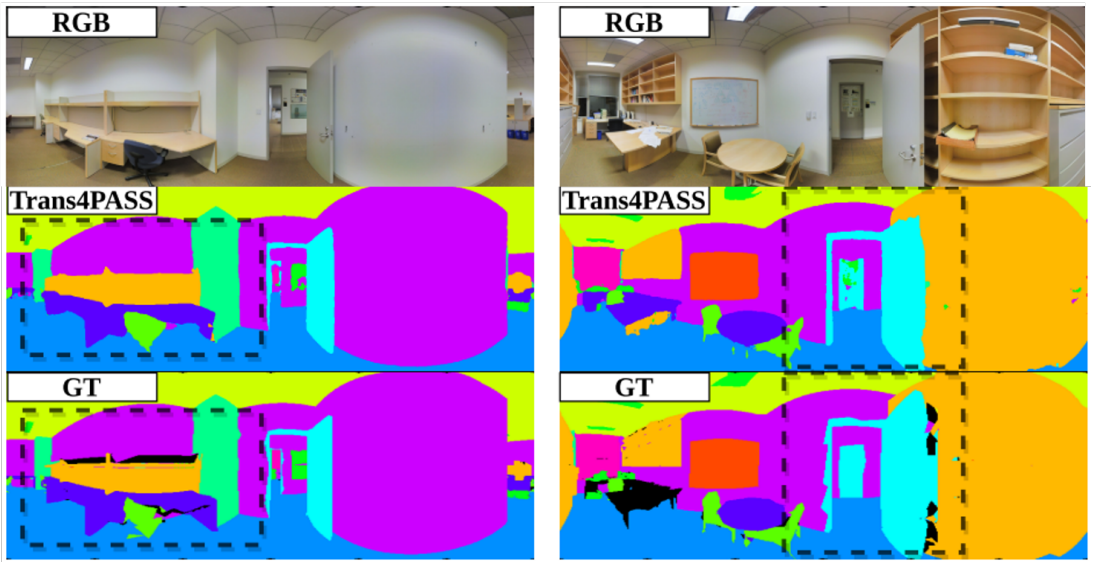

Fig. 4: Zhang *et al.* [14]

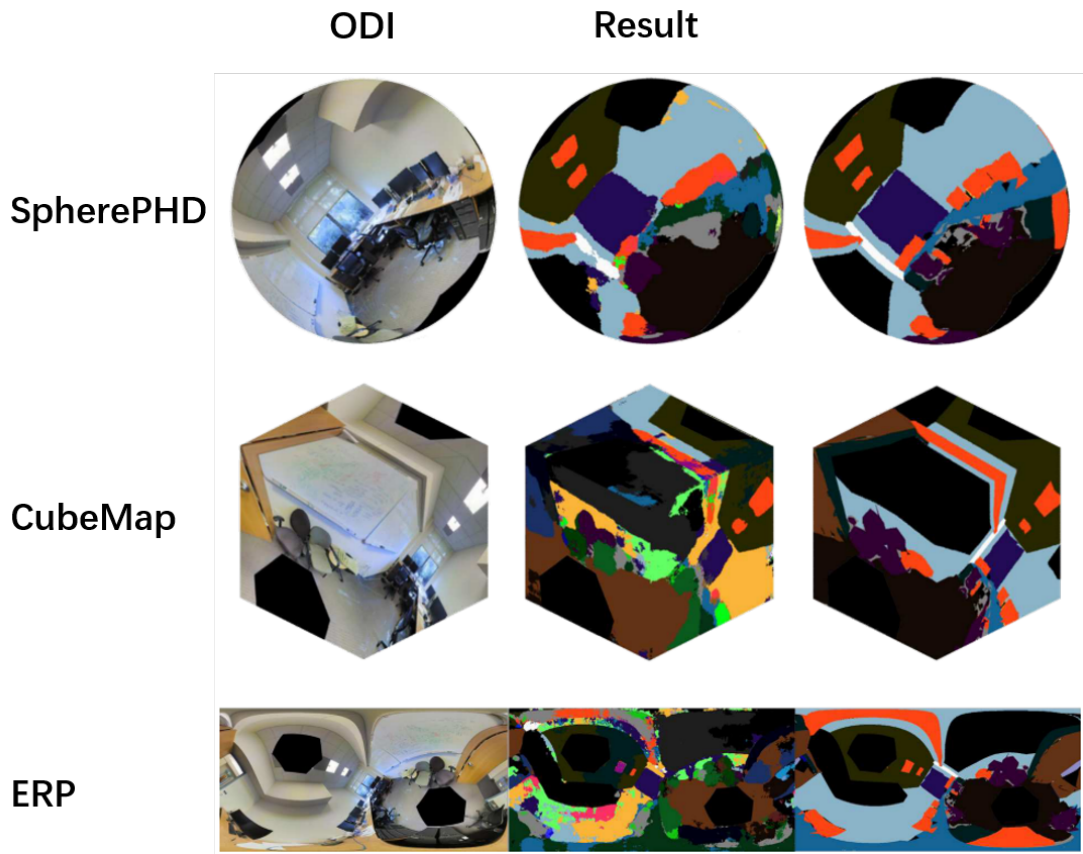

Fig. 5: Lee *et al.* [15]

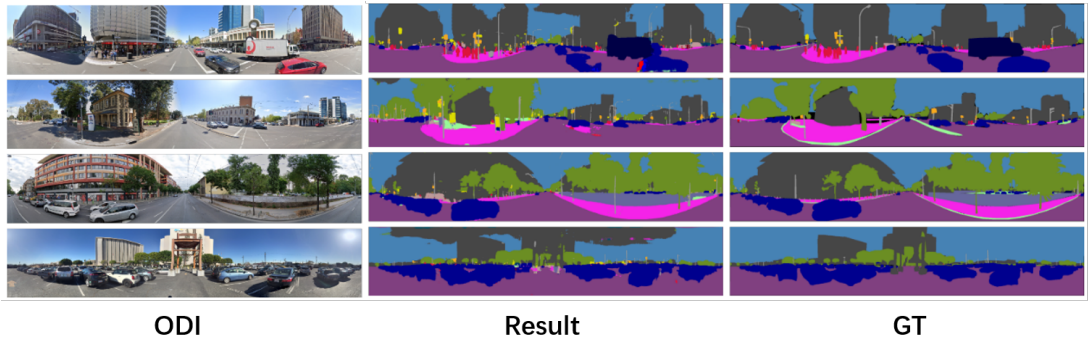

Fig. 6: Ma *et al*. [17]

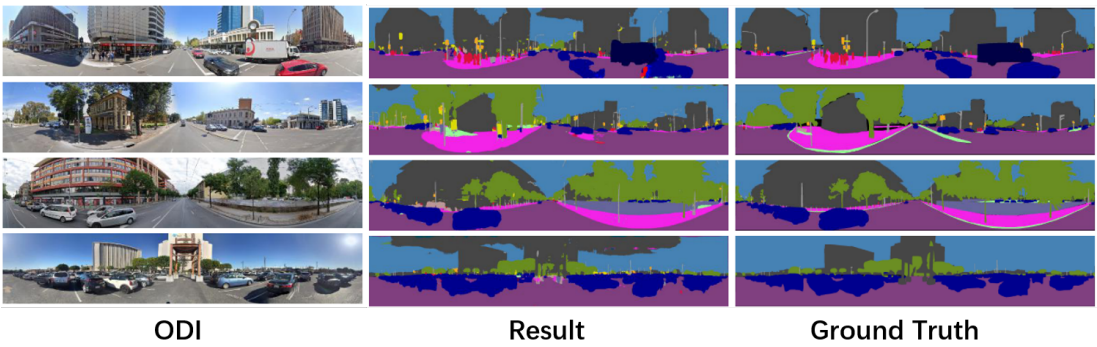

Fig. 7: Zhang *et al*. [18]

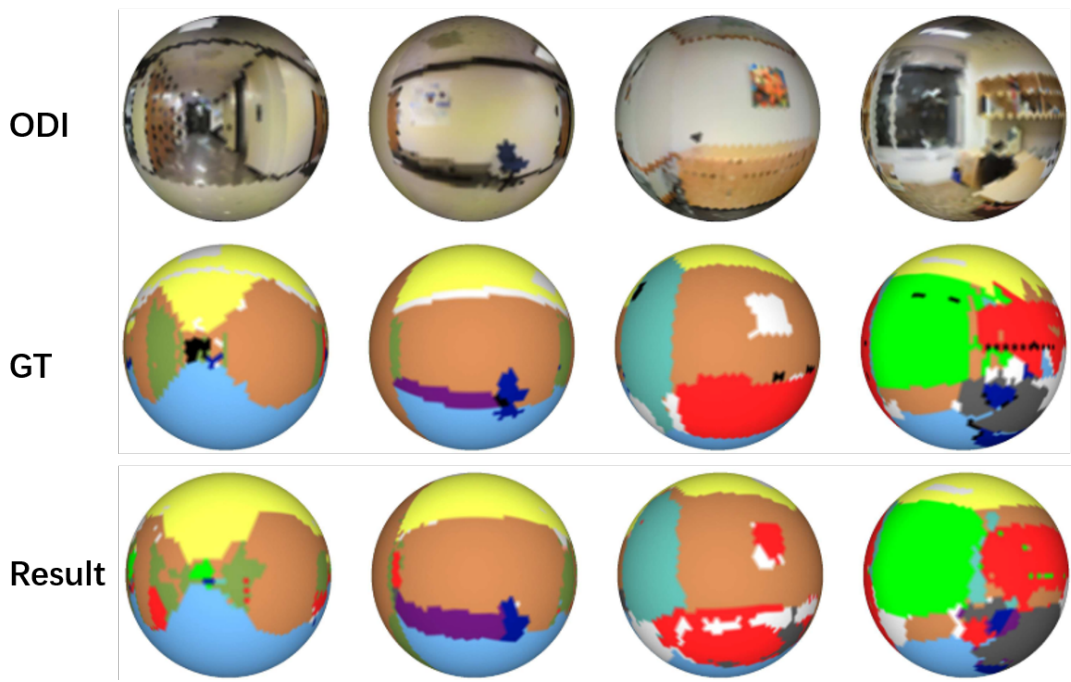

Fig. 8: Zhang *et al*. [16]

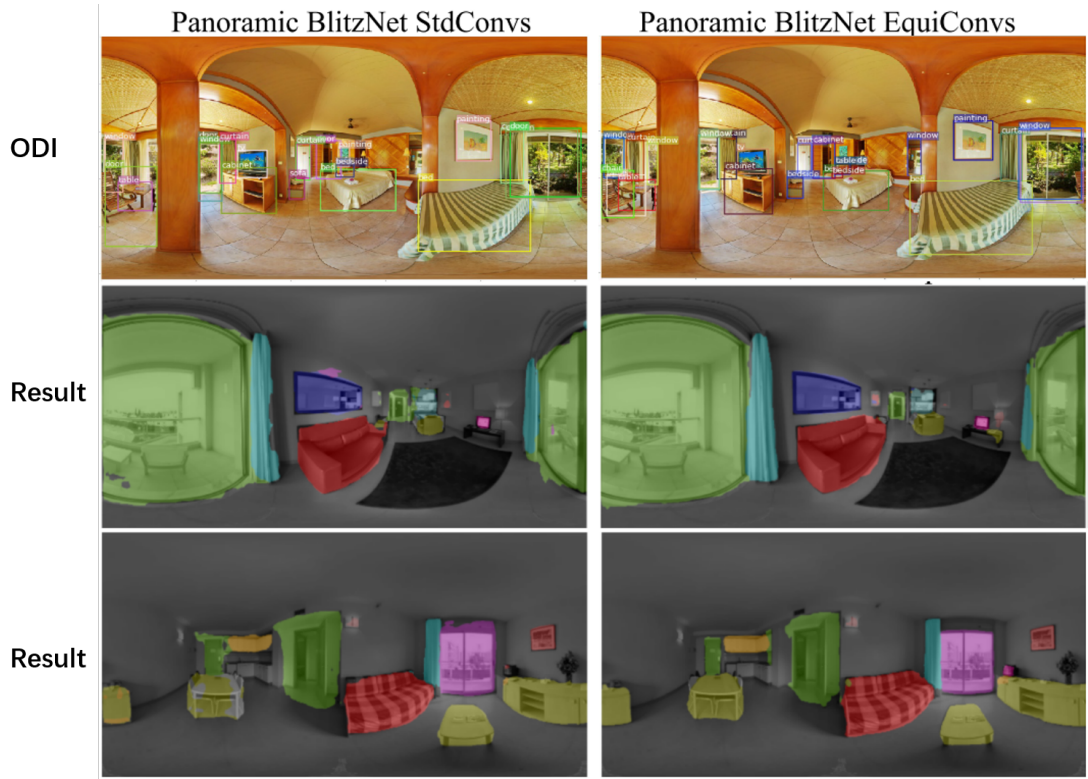

Fig. 9: Viu *et al.* [76]

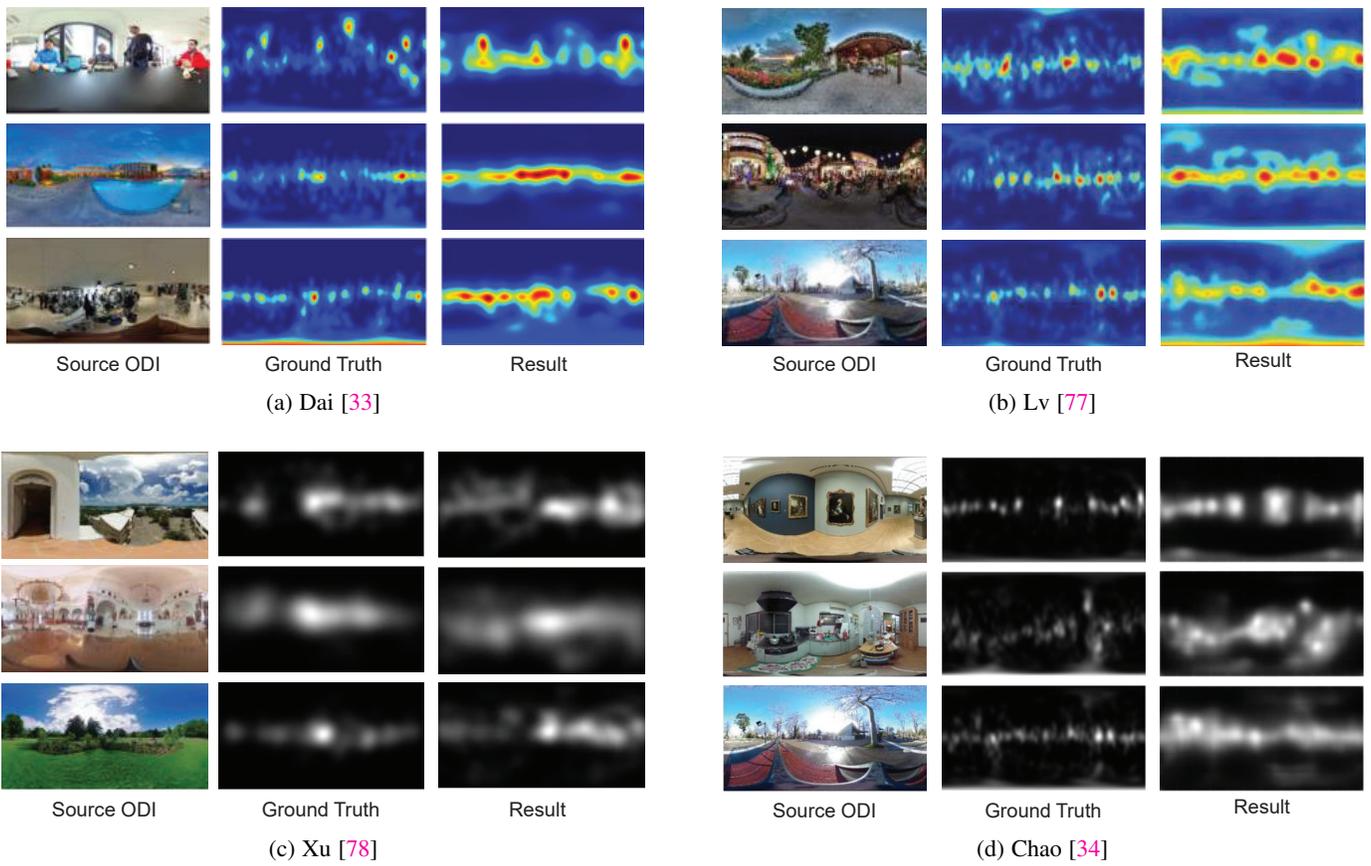

(a) Dai [33]

(b) Lv [77]

(c) Xu [78]

(d) Chao [34]

Fig. 10: Visual results on the representative ODI saliency prediction methods.

TABLE 6: Qualitative comparison of ODV saliency prediction.

| Dataset | Method | KLD ↓ | CC ↑ | NSS ↑ | AUC ↑ | sAUC ↑ | AUC_J ↑ | AUC_B ↑ |
|---|---|---|---|---|---|---|---|---|
| Panoramic Saliency [79] | Nguyen [79] | - | 0.2521 | 1.9864 | - | 0.7112 | - | - |
| Wild-360 [58] | Chen [58] | - | 0.42 | - | - | - | 0.898 | 0.859 |
| Salient360 [32] | Zhang [80] | - | 0.4087 | 0.6989 | - | - | 0.6594 | - |
| DOM [81] | Zhang [80] | 5.724 | 0.533 | 2.502 | - | - | 0.882 | - |
| PVS-HM [82] | Xu [82] | - | 0.185 | 0.899 | - | 0.72 | - | - |
|  | Qiao [83] | 1.253 | 0.585 | 2.627 | - | - | - | - |


\fi

\end{document}